\documentclass[11pt,a4paper]{article}

\usepackage{polyu-vclab}

\usepackage{lipsum}                
\usepackage{amsmath}
\usepackage{booktabs}
\usepackage{multirow}
\usepackage{graphicx}
\usepackage{booktabs}
\usepackage{algorithm}
\usepackage{algorithmic}
\usepackage{adjustbox}
\usepackage{afterpage}
\usepackage{float}
\usepackage{multirow}   
\usepackage{xcolor}
\usepackage[table]{xcolor} 
\usepackage{tabularx}   
\usepackage{graphicx}   
\usepackage{bbm}
\usepackage{makecell}
\usepackage{subcaption} 
\usepackage{amsmath}
\usepackage{dsfont}
\usepackage{pifont}
\usepackage{hyperref}
\usepackage{cleveref}
\usepackage[numbers,sort&compress]{natbib}

\newcommand{\cmark}{{\color{green!70!black}\ding{51}}} 
\newcommand{\xmark}{{\color{red!70!black}\ding{55}}} 
\crefname{figure}{Fig.}{Figs.}
\crefname{table}{Tab.}{Tabs.}

\setEyebrow{PolyU VCLab\,\textbullet\,ECCV 2026}

\setReportTag{Visual Computing Lab\,\textperiodcentered\,The Hong Kong Polytechnic University}

\papertitle{Dual Distribution Estimation for Zero-shot Noisy Test-Time Adaptation with VLMs}

\paperauthors{%
  Wenjie Zhu\affilmark{1}\quad
  Yabin Zhang\affilmark{3}\quad
  Liang Xu\affilmark{4}\quad
  Xin Jin\affilmark{2}\quad
  Wenjun Zeng\correspondmark\affilmark{2}
  Lei Zhang\correspondmark\affilmark{1}%
}

\paperaffil{%
  \affilmark{1}\,The Hong Kong Polytechnic University\quad
  \affilmark{2}\,Eastern Institute of Technology, Ningbo\quad \\
  \affilmark{3}\,Harbin Institute of Technology (Shenzhen)\quad
  \affilmark{4}\,Shanghai Jiao Tong University\quad
}

\papernotes{%
  \correspondmark\,Corresponding author (\href{mailto:cslzhang@comp.polyu.edu.hk}{cslzhang@comp.polyu.edu.hk}).%
}

\paperbadges{%
  \vclabbadgesolid{Project Page}{https://zhuwenjie98.github.io/DDE-project-page/}\;%
  \vclabbadgesolid{Code}{https://github.com/PolyU-VCLab/openood-vlm}\;%
}

\begin{document}

\maketitleVCLab

\begin{vclabAbstract}
\noindent\textbf{Abstract.}\;
While test-time adaptation (TTA) empowers vision-language models to adapt without costly retraining, it remains highly vulnerable to out-of-distribution (OOD) outliers prevalent in real-world applications. This discrepancy motivates Noisy TTA (NTTA), an online task to filter noisy OOD samples on the fly while maximizing in-distribution (ID) classification accuracy. Existing zero-shot NTTA approaches typically rely on test-time discriminative training, leading to overconfident misclassifications and significantly degraded inference efficiency. To address these limitations, we propose a novel framework named Dual Distribution Estimation (DDE), shifting the zero-shot NTTA paradigm from instance-level learning to training-free Gaussian distribution modeling. DDE incorporates two novel modules: Positive Feature Distribution Estimation (PFDE) and Negative Label Distribution Estimation (NLDE). PFDE explicitly models class-wise inclusion and exclusion Gaussian distributions to formulate a calibrated contrastive score, robustly enhancing ID accuracy. In parallel, NLDE improves OOD identification by explicitly modeling the negative label distribution to mine highly discriminative labels, effectively mitigating spurious correlations. Extensive experiments show that on the large-scale ImageNet benchmark, DDE achieves an improvement of 3.70\% in harmonic mean accuracy and reduces the FPR95 for OOD detection by 6.20\%, while ensuring highly scalable and efficient online inference. Furthermore, DDE is zero-shot and training-free, demonstrating remarkable robustness in data-scarce scenarios. Codes are available at \url{https://github.com/PolyU-VCLab/OpenOOD-VLM}.
\end{vclabAbstract}


\section{Introduction}

\begin{figure}[t]
    \centering
    \begin{subfigure}[t]{0.31\linewidth}
        \centering
        \includegraphics[width=\linewidth]{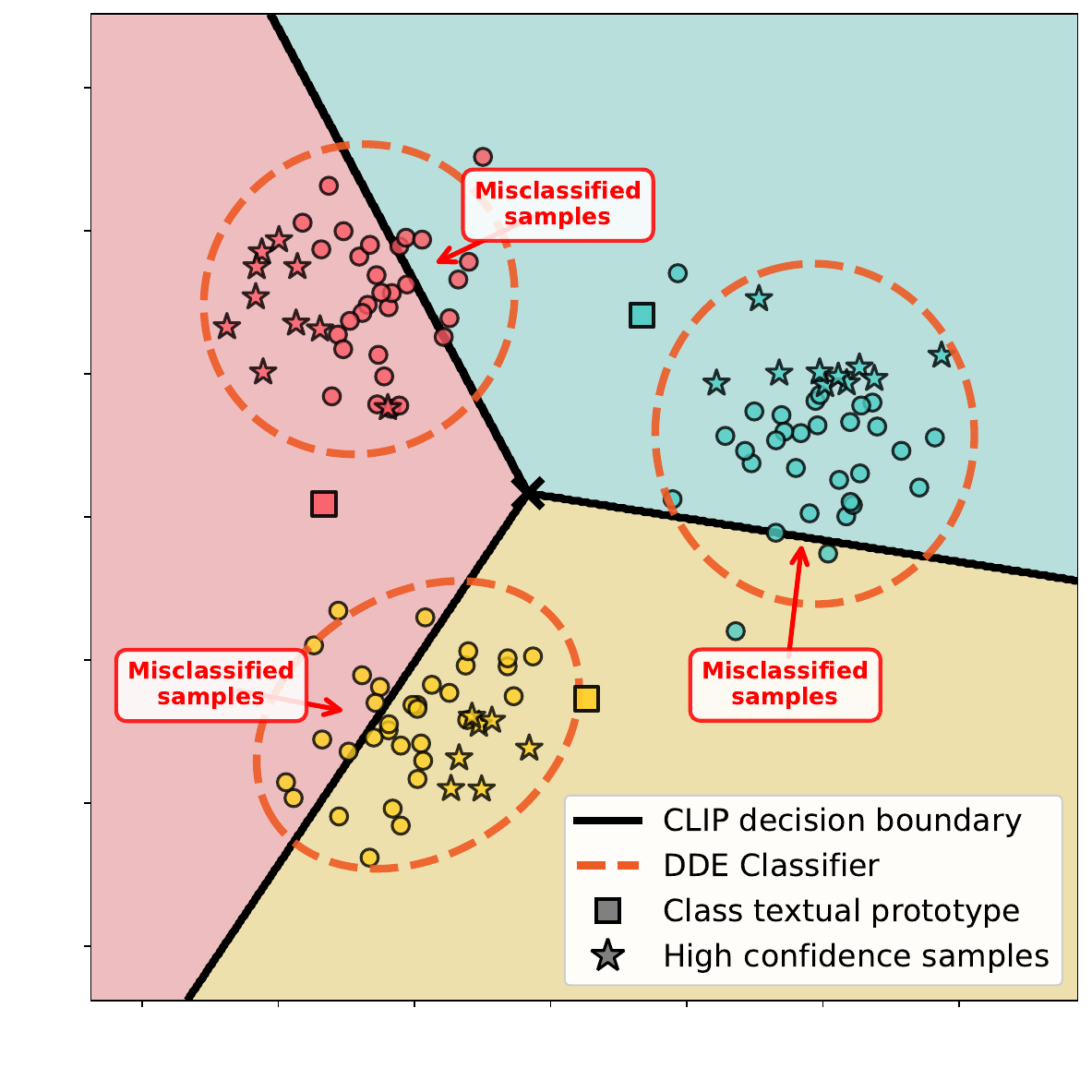}
\caption{CLIP vs. DDE Classifiers}
        \label{fig:gda_classifier}
    \end{subfigure}
    \hfill
    \begin{subfigure}[t]{0.31\linewidth}
        \centering
        \includegraphics[width=\linewidth]{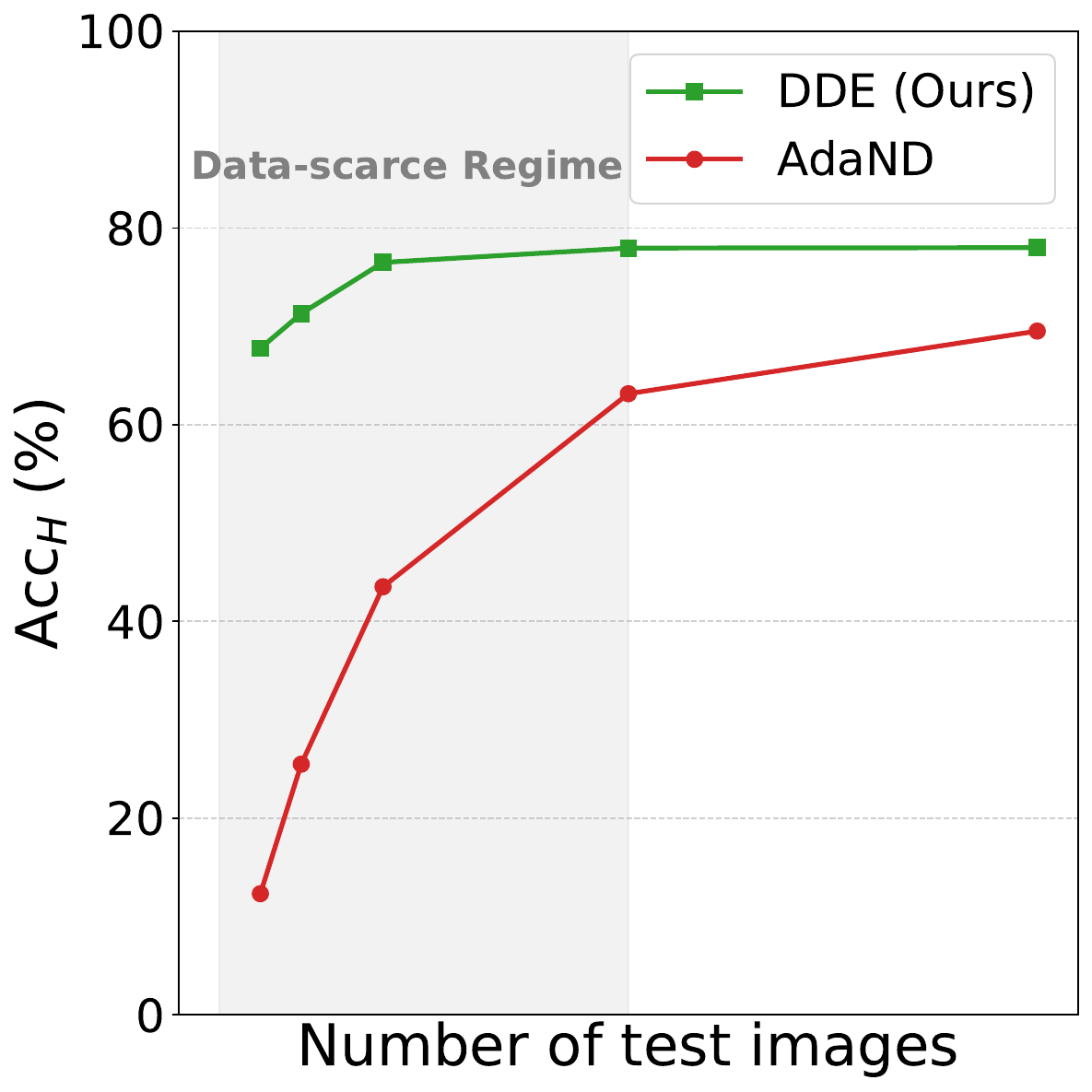}
        \caption{Data scarce setting}
        \label{fig:data_scarce}
    \end{subfigure}
    \hfill  
    \begin{subfigure}[t]{0.31\linewidth}
        \centering
        \includegraphics[width=\linewidth]{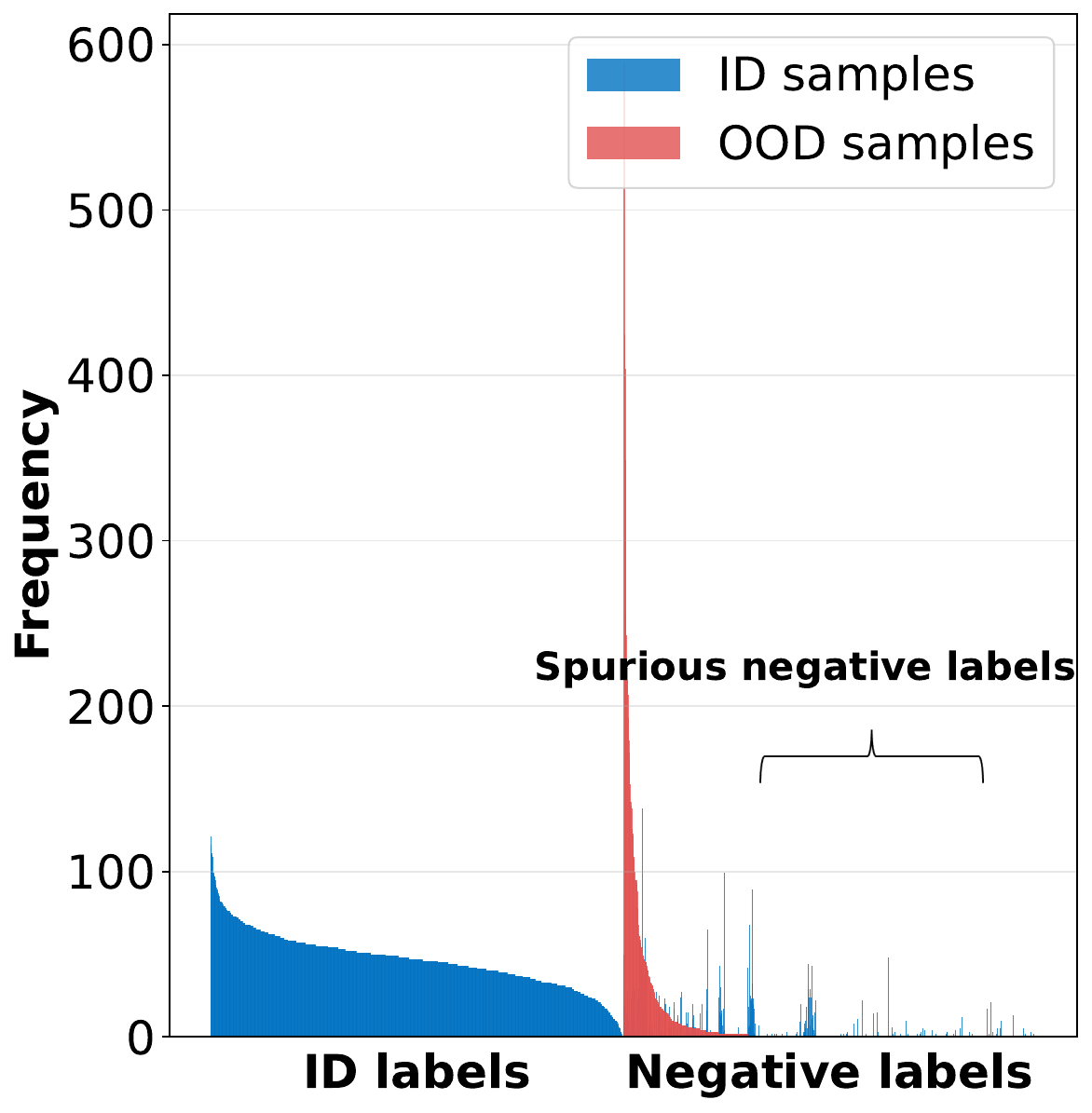}
        \caption{Label distribution}
        \label{fig:neglabel_bias}
    \end{subfigure}
\caption{\textbf{(a) CLIP vs. DDE Classifiers:} AdaND~\cite{cao2025noisy} uses the CLIP classifier frequently misclassifying samples that deviate significantly from textual prototypes. Other TTA methods~\cite{zhang2024dual_cvpr} (e.g., DMN) cache high-confidence samples, which still struggle to accurately represent complete visual class distribution. In contrast, our DDE achieves a more precise visual representation by effectively modeling the test distribution. \textbf{(b) Data scarce setting:} Unlike AdaND, which relies on extensive ID and noisy images for training, DDE dynamically models discriminative negative labels. This allows DDE to maintain higher robustness in data-scarce environments.  \textbf{(c) Label distribution:} Some negative labels exhibit spurious correlations with clean ID data, whereas OOD outliers strongly align with only a narrow subset of negative labels.}
    \label{fig:comparison}
\end{figure}

Test-time adaptation (TTA) for vision-language models (VLMs) has garnered increasing attention, as it enhances pre-trained models directly during deployment without requiring costly human annotation or retraining. Most existing TTA methods operate under a closed-world assumption that all test data strictly fall within a specific in-distribution (ID) label space~\cite{zhang2024dual_cvpr,karmanov2024efficient,chen2025multi, zhang2024dual_neurips}. However, real-world test streams inevitably contain out-of-distribution (OOD) outliers. This discrepancy motivates the emerging task of Noisy TTA (NTTA)~\cite{cao2025noisy}, which operates under a strict online setting with dual objectives: simultaneously filtering out noisy OOD samples on the fly and maximizing classification accuracy on clean ID data. Crucially, this dual-objective, online nature distinguishes NTTA from TTA for OOD detection~\cite{zhang2024adaneg, yang2025oodd}, which typically focuses solely on offline outlier evaluation with metrics like AUROC.

NTTA methodologies generally fall into two categories: those requiring task-specific training \cite{li2023robustness}, and zero-shot NTTA, which leverages frozen VLMs for task-agnostic inference \cite{cao2025noisy}. In this work, we focus on the more challenging and versatile zero-shot NTTA setting. A representative approach in this domain is AdaND \cite{cao2025noisy}, which trains a noise detector online using selected test samples. However, several limitations persist in this approach: (1) AdaND does not model test distribution historical test image features, relying instead on the CLIP text-image alignment for classification~(\cref{fig:gda_classifier}). This leads to frequent misclassifications when samples deviate significantly from the class textual prototypes. (2) AdaND necessitates a substantial number of samples to achieve satisfactory performance, which limits the robustness of the system in data-scarce scenarios~(\cref{fig:data_scarce}). (3) AdaND relies on online training, which significantly degrades inference efficiency~(\cref{tab:complexity}). These limitations are summarized in~\cref{tab:method_comparison}.

To address these limitations, we propose Dual Distribution Estimation (DDE), which is empowered by two novel modules: Positive Feature Distribution Estimation (PFDE) and Negative Label Distribution Estimation (NLDE). Specifically, PFDE enhances ID sample recognition by modeling mined positive image features as class-conditional Gaussian distributions and employing Bayes' theorem for robust posterior inference. To capture a more precise decision boundary, we introduce a dual Gaussian modeling approach that explicitly characterizes both the inclusion and exclusion distributions for each class. The final prediction is formulated as a calibrated contrastive score by subtracting the exclusion probability from the inclusion one.

\begin{table}[t]\centering
\caption{Comparison between methods of TTA, TTOOD (Test time OOD Detection), NTTA and our DDE. DDE estimates dual-distributions from historical image features, uniquely combining the ability to handle noisy data with a robust, training-free, high-efficiency architecture.}
\label{tab:method_comparison}
\resizebox{\textwidth}{!}{%
\begin{tabular}{@{}lcccc@{}}
\toprule
\textbf{Methods} & \textbf{DMN}~\cite{zhang2024dual_cvpr} & \textbf{AdaNeg}~\cite{zhang2024adaneg} & \textbf{AdaND}~\cite{cao2025noisy} & \textbf{DDE}  \\
 & (TTA) & (TTOOD) & (NTTA) & (NTTA) \\ \midrule
OOD generalization & \cmark & \xmark & \cmark & \cmark \\
OOD(Noisy) detection & \xmark & \cmark & \cmark & \cmark \\
Metrics & Accuracy & AUROC, FPR95 & $Acc_{S}$, $Acc_{N}$, $Acc_{H}$ & $Acc_{S}$, $Acc_{N}$, $Acc_{H}$ \\ \midrule
\textbf{Distribution Modeling} & \xmark & \xmark & \xmark & \cmark \\
Robustness to data-scarce scenarios & \cmark & \cmark & \xmark & \cmark \\
Training-free & \cmark & \cmark & \xmark & \cmark \\
Fast inference speed & \cmark & \cmark & \xmark & \cmark \\ \bottomrule
\end{tabular}%
}
\end{table}

In parallel, NLDE enhances noisy OOD perception by mining highly discriminative negative labels. This module is motivated by a key observation in \cref{fig:neglabel_bias} that some negative labels exhibit spurious correlations with clean ID data and true OOD outliers strongly align with only a narrow subset of specific negative labels. To exploit this, we model the image-to-label distribution and mine the most discriminative negative labels from the corpora dataset by quantifying their similarity discrepancy across positive and negative images. Finally, we integrate an adaptive thresholding mechanism \cite{li2023robustness} to dynamically separate ID and OOD samples during the online inference process.

We conduct extensive experiments to validate the advantages of DDE. On the large-scale ImageNet dataset, our DDE achieves a significant 3.70\% improvement in harmonic mean accuracy while reducing FPR95 for OOD detection by 6.20\%. Moreover, our method operates in a zero-shot and training-free manner, demonstrating strong scalability. We summarize our contributions as follows:
\begin{itemize}
    \item We first identify several critical limitations of the existing AdaND~\cite{cao2025noisy} method in the zero-shot Noisy TTA setting: (1) it neglects historical test image feature distribution, thereby failing to capture class-specific diversity; (2) it lacks robustness in data-scarce scenarios, as it relies on extensive samples to train its noise detector; and (3) it requires online training, which significantly degrades inference efficiency.
    \item To address these pitfalls, we propose a novel framework named Dual Distribution Estimation, which consists of two novel, complementary modules. The Positive Feature Distribution Estimation formulates a calibrated contrastive score via class-wise inclusion and exclusion Gaussian distributions to enhance ID accuracy. In parallel, the Negative Label Distribution Estimation improves OOD identification by mining highly discriminative negative labels, effectively isolating true OOD outliers and mitigating spurious similarities among generic negative labels.
    \item Extensive experiments on large-scale benchmarks validate the superiority of our approach. On ImageNet, DDE achieves a remarkable 3.70\% improvement in harmonic mean accuracy and reduces the FPR95 for OOD detection by 6.20\%. Crucially, as a fully zero-shot and training-free method, DDE ensures highly scalable and efficient online inference.
\end{itemize}

\section{Related Work}
\noindent\textbf{Test-time Adaptation.} 
TTA~\cite{wang2020tent,niu2023towards} allows pre-trained models to adjust to unlabeled data during deployment. By updating parameters or statistics on-the-fly, TTA addresses distribution shifts between training and real-world environments without requiring access to the source data. Existing approaches often rely on self-supervised objectives~\cite{shi2022efficacy, meng2026topocl, ma2025instruct}, such as entropy minimization~\cite{wang2020tent, niu2023towards} or recalibrate batch normalization statistics, to align feature distributions~\cite{mirza2022norm, zhao2023delta}. Recent studies~\cite{zhang2024dual_cvpr, karmanov2024efficient, sui2025just, zhang2024dual_neurips, chen2025multi, zhu2024enhancing, han2024dota, lafon2025cliptta, zhou2025bayesian, luo2026protodcs, zhang2025model} have extended these techniques to VLMs. Many of these VLM-based methods~\cite{sui2025just, zhang2024dual_neurips} employ test-time prompt tuning or residual learning to derive adaptive representations for individual samples. Alternatively, some frameworks~\cite{zhang2024dual_cvpr, karmanov2024efficient} investigate training-free mechanisms, such as dynamic adapters, to improve efficiency. Other strategies~\cite{zhu2024enhancing, han2024dota} involve learning prompt distributions to account for diverse visual features. Despite these advancements, current methods often fail to account for noisy samples in the data stream, which can lead to substantial performance degradation.

\noindent\textbf{Robustness under Distribution Shift.}
Recent research has studied model robustness under noisy and distribution-shifted. Recent works~\cite{yang2023re, yang2024backdoor, lu2025out, feng2026beyond, feng2026seeing, feng2026generalized, ning2024physics, ning2025physics, he2026cinematte,zhu2024advancing, zhu2026medeyes, tao2026grasp, fu2026modalimmune, wang2026sppo, yang2024regulating} enhance model interpretation by refining feature robustness. OWTTT~\cite{li2023robustness} introduced the Open-World Test-Time Adaptation benchmark and utilized adaptive thresholding to exclude OOD samples. However, it relies on source data and is therefore unsuitable for source-free VLMs such as CLIP. AdaND \cite{cao2025noisy} addresses zero-shot noisy TTA but depends on pseudo-labels to train a noise detector, which is ineffective in data-scarce settings.

\noindent\textbf{Test-time OOD Detection on Vision Language Models.} The cross-modal alignment of VLMs has driven progress in OOD detection, with recent work~\cite{zhang2024adaneg, yang2025oodd, zhang2024lapt, zhu2025knowledge, zhu2026ants, zhang2026activation, zhu2025knowledge, zhang2026activation, tang2026cross, guo2025quantized,zeng2025FSDrive, xiao2026reversible, zhang2026pi, ma2026thinking, li2025mmt, li2025ammkd} shifting toward test-time paradigms that adjust to the inference stream. AdaNeg~\cite{zhang2024adaneg} and OODD~\cite{yang2025oodd} employ memory banks and dynamic dictionaries to maintain adaptive negative proxies, thereby addressing the semantic gaps found in static labels. Nevertheless, these strategies frequently rely on high-confidence samples, which can prevent them from fully capturing the underlying data distribution.

\begin{figure}[t]
  \centering
  \includegraphics[width=\linewidth]{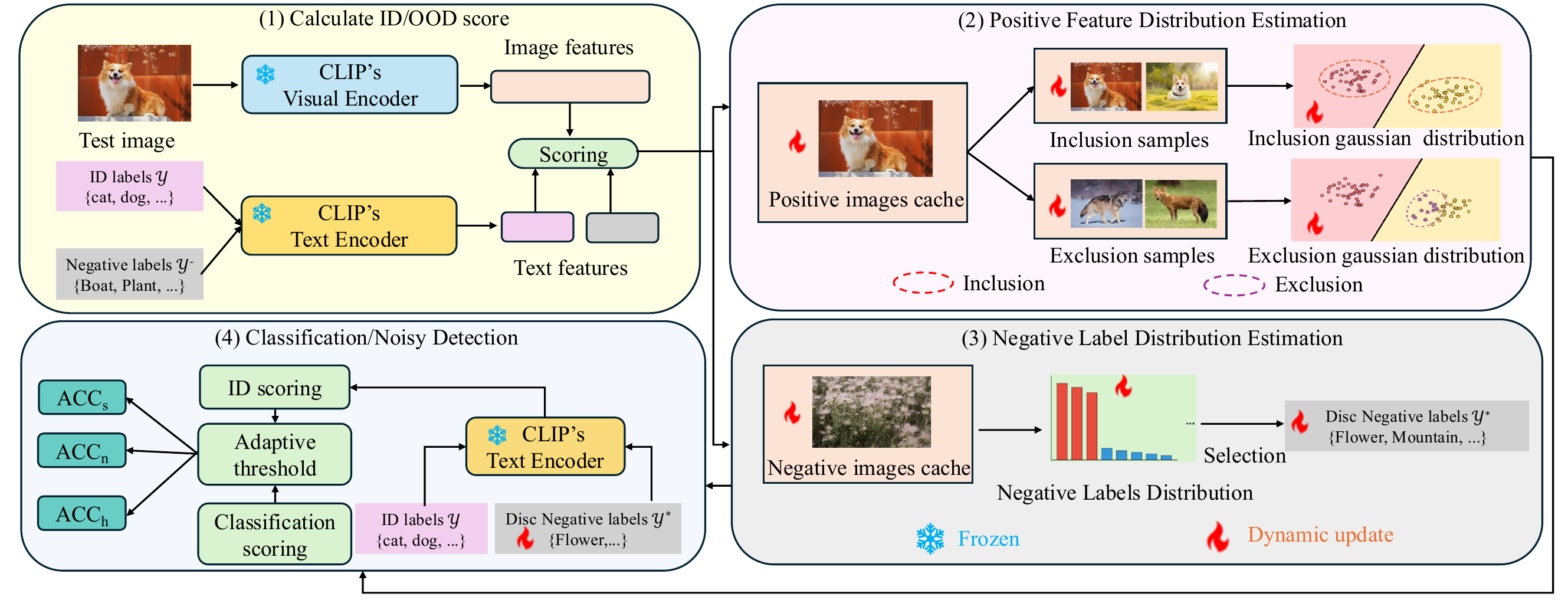}
  \caption{Overview of our DDE framework. (1) Calculate the ID/OOD score for test images, dynamically caches positive and negative samples. (2) Model positive feature Gaussian distributions via inclusion and exclusion GDA estimating. (3) Filter the discriminative negative labels via negative labels distribution estimation. (4) Employ an adaptive threshold for simultaneous ID classification and noise detection.}
  \label{fig:framework}
  \vspace{-0.1cm}
\end{figure}

\section{Method}
\subsection{Preliminaries}
\noindent\textbf{Gaussian Discriminant Analysis.}
Inspired by \cite{hastie1996discriminant,han2024dota}, we assume that the embedding distribution of each class $k$ follows a Gaussian distribution as $P(x \mid y = k) \sim \mathcal{N}(\mu_k, \Sigma_k)$, where $\mu_k$ and $\Sigma_k$ are the mean vector and covariance matrix of class $k$, respectively.
For the $k$-th class with $N_k$ training samples, the mean and covariance matrix of different classes can be estimated as follows:
\begin{equation} \label{eq:gda}
\mu_k = \frac{\sum_{n=1}^{N_k} x_n}{N_k},
\qquad
\Sigma_k =
\frac{\sum_{n=1}^{N_k} (x_n - \mu_k)(x_n - \mu_k)^{T}}
{N_k},
\end{equation}
where $\mu_k$ and $\Sigma_k$ represent the estimated mean and covariance matrix of the $k$-th class, respectively. $x_n$ denotes the input embeddings. 

\noindent\textbf{OOD Detection with Negative Labels.}  
Enhancing OOD detection with textual knowledge has recently gained traction \cite{ming2022delving,zhang2024adaneg}, particularly through the use of negative labels \cite{jiang2024negative}. In addition to standard ID labels $\mathcal{Y}$, these methods typically introduce a disjoint set of negative labels $\mathcal{Y}^-$ and classify test samples as OOD when they exhibit higher similarity to negative labels than to ID classes. Since the performance of this approach relies on the relevance of the negative set, NegLabel \cite{jiang2024negative} constructs the set $\mathcal{Y}^-$ by selecting the top $M$ labels from a large corpus $\mathcal{Y}^{cor} = \left\{ \widetilde{y}_1, \widetilde{y}_2, \ldots, \widetilde{y}_{N_c} \right\}$ that exhibit the most significant cosine distance from the ID labels. Formally, this selection is defined as $\mathcal{Y}^{-} = \operatorname{Top} \left( \{ d_i \}_{i=1}^{{N_c}}, \, \mathcal{Y}^{cor}, \, M \right)$, where $N_{c}$ and $M$ denote the number of candidate labels in the corpus and the number of selected negative labels, respectively. The scoring function for OOD detection is defined as:
\begin{equation} \label{eq:neglabel_score}
S_{\text{nl}}(x) = \sum_{i=1}^{K} \frac{\exp(x^\top \mathbf{w}_{i} / \tau)}{\sum_{j=1}^{K} \exp(x^\top \mathbf{w}_{j} / \tau) + \sum_{j=1}^{M} \exp(x^\top \tilde{\mathbf{w}}_j / \tau)}.
\end{equation}
In this formulation, $x \in \mathcal{R}^D$ denotes the test image feature, while $\mathbf{w}_i \in \mathcal{R}^D$ and $\tilde{\mathbf{w}}_j \in \mathcal{R}^D$ are the textual prototypes for the ID labels $y_i \in \mathcal{Y}$ and negative labels $y^-_j \in \mathcal{Y}^-$, respectively, with $D$ being the feature dimension. Meanwhile, $\tilde{\mathbf{w}}_j = f_{\text{txt}}(\rho(\tilde{y}_j)) \in \mathcal{R}^D$ represents the text prototype of the mined negative label $\tilde{y}_j$, and $\tau > 0$ is the temperature parameter.

\subsection{Motivation}
Current zero-shot NTTA methods, such as AdaND~\cite{cao2025noisy}, often rely on simple linear classifiers for inference and noise detection. However, these classifiers fail to account for the underlying test distribution, thereby neglecting the diverse image features within each category. Furthermore, the absence of OOD-aware information aggregation results in redundant negative labels, which introduces interference and undermines classification accuracy. These challenges motivate the following research question: \textit{Can we design a method to accurately estimate both the positive feature distribution and the negative label distribution?}

In this work, we develop a dual distribution estimation framework, named DDE, as illustrated in~\cref{fig:framework}. These two distribution estimation components to facilitate ID classification and OOD detection, respectively, ultimately enhancing the performance of NTTA.

\subsection{Dual Distribution Estimation}
\noindent\textbf{Positive and Negative Images Selection.}
Following \cref{eq:neglabel_score}, we identify positive and negative samples for each batch by applying predefined thresholds $\lambda_{pos}$ and $\lambda_{neg}$. Specifically, we follow AdaNeg~\cite{zhang2024adaneg}
defining the positive set as $\mathcal{X}_{pos}^{batch} = \{ v \in \mathcal{R}^D \mid S_{nl}(v) \geq \lambda_{pos} \}$ and the negative set as $\mathcal{X}_{neg}^{batch} = \{ v \in \mathcal{R}^D \mid S_{nl}(v) < \lambda_{neg} \}$, where $\lambda_{pos}$ and $\lambda_{neg}$ represent the thresholds for isolating high-confidence pseudo-positive and pseudo-negative samples. Once the positive images $\mathcal{X}_{pos}^{batch}$ and negative images $\mathcal{X}_{neg}^{batch}$ are identified, they are stored in two separate caches, $\mathcal{C}_{pos}$ and $\mathcal{C}_{neg}$ as:
\begin{equation} \label{eq:pos_neg_images_cache}
\begin{aligned}
\mathcal{C}_{pos} &\leftarrow \text{Concat}(\mathcal{C}_{pos}, \mathcal{X}_{pos}^{batch}) \quad \text{s.t.} \quad |\mathcal{C}_{pos}| \leq Q, \\
\mathcal{C}_{neg} &\leftarrow \text{Concat}(\mathcal{C}_{neg}, \mathcal{X}_{neg}^{batch}) \quad \text{s.t.} \quad |\mathcal{C}_{neg}| \leq Q.
\end{aligned}
\end{equation}
$Q$ indicates the maximum capacity constraints for each cache. Then we use them to estimate the positive features distribution and the negative labels distribution.\\

\noindent\textbf{Positive Feature Distribution Estimation}.
\noindent\textit{(1) Gaussian Discriminant Analysis.}
Following the generative framework of Gaussian Discriminant Analysis (GDA) \cite{bishop2006pattern}, we adopt a dual Gaussian modeling approach. In this setting, the classifier is derived from specific assumptions about the data distribution for each class. 
Following the standard GDA framework, the features are assumed to follow Gaussian distributions with a class-conditional covariance matrix, i.e., $P(x|y=k) \sim \mathcal{N}(\mu_k, \Sigma_k)$. Then we use the logit function defined in GDA~\cite{bishop2006pattern}, which measures how well a sample $x$ fits the distribution of class $k$ as:
\begin{equation} \label{eq:gda_logits}
f_k(x) = \mu_k^T \Sigma^{-1}_{k} x - \frac{1}{2} \mu_k^T \Sigma^{-1}_{k} \mu_k + \log p_k.
\end{equation}
Based on the Bayes' theorem, the posterior probability $P(y = k \mid x)$ for a given sample $x \in \mathcal{X}_{pos}^{batch}$ is determined by combining the class-conditional density $P(x \mid y = k)$ with a uniform prior $P(y = k) = 1/K$:
\begin{equation} \label{eq:gda_classifier}
P(y = k \mid x) = \frac{P(x \mid y = k) P(y = k)}{\sum_{j=1}^{K} P(x \mid y = j) P(y = j)} = \frac{\exp(f_k(x))}{\sum_{j=1}^{K} \exp(f_j(x))}.
\end{equation}

\noindent\textit{(2) Partition the Positive Images.}
Unlike prior works~\cite{zhu2024enhancing, han2024dota} that aggregate all soft assignments into a single distribution, we explicitly disentangle the positive feature space into inclusion and exclusion distributions. For each ID class $k$, we maintain two Gaussian distributions, $\mathcal{N}(\boldsymbol{\mu}_k^{\mathrm{in}}, \boldsymbol{\Sigma}_k^{\mathrm{in}})$ and $\mathcal{N}(\boldsymbol{\mu}_k^{\mathrm{ex}}, \boldsymbol{\Sigma}_k^{\mathrm{ex}})$, modeled from inclusion and exclusion features, respectively. Both distributions are updated via test-time streaming. At each test-time batch, PFDE is updated in a class-wise manner using the positive samples $\mathcal{X}_{pos}^{batch}$ selected from the current batch.
Specifically, for a given class $k$, the inclusion distribution 
$\mathcal{N}(\boldsymbol{\mu}_k^{\mathrm{in}}, \boldsymbol{\Sigma}_k^{\mathrm{in}})$ 
is updated using features of samples in $\mathcal{X}_{pos}^{batch}$ whose top-1 CLIP prediction is class $k$. 
The exclusion distribution $\mathcal{N}(\boldsymbol{\mu}_k^{\mathrm{ex}}, \boldsymbol{\Sigma}_k^{\mathrm{ex}})$ 
is updated using features of samples in $\mathcal{X}_{pos}^{batch}$ whose top-1 prediction is not $k$ but whose second-highest CLIP prediction corresponds to class $k$. These samples represent near-boundary cases that are easily confused with class $k$, allowing the exclusion distribution to characterize ambiguous regions in the feature space.

\noindent\textit{(3) Estimating Gaussian Distribution Parameters.}
Given a batch of positive test samples $\mathcal{X}_{pos}^{batch}=\{x_n\}_{n=1}^{N^{batch}}$ at time step $t$, we estimate the class-specific parameters for each category $k$, namely the means $\{\boldsymbol{\mu}_k^{\mathrm{in}}, \boldsymbol{\mu}_k^{\mathrm{ex}}\}_{k=1}^K$ and covariances $\{\boldsymbol{\Sigma}_k^{\mathrm{in}}, \boldsymbol{\Sigma}_k^{\mathrm{ex}}\}_{k=1}^K$. 
We estimate the parameters via the Expectation-Maximization (EM) algorithm. Specifically, the expectation step derives zero-shot predictions, while the maximization step refines the model parameters by maximizing the log-likelihood of the observed data as:
\begin{equation} \label{eq:gauss_dis_estimation}
\small
\begin{aligned}
    \boldsymbol{\mu}_{k, t}^{s} &= \frac{N_{k, t-1}^{s} \boldsymbol{\mu}_{k, t-1}^{s} + \sum_{i=1}^{N^{batch}} P_k(y = k \mid x_i) x_i }{N_{k, t}^{s} + \sum_{i=1}^{N^{batch}} P_k(y = k \mid x_i)}, \\
    \boldsymbol{\Sigma}_{k, t}^{s} &= \frac{N_{k, t-1}^{s} \boldsymbol{\Sigma}_{k, t-1}^{s} + \sum_{i=1}^{N^{batch}} P_k(y = k \mid x_i) (x_i - \boldsymbol{\mu}_{k, t}^{s})(x_i - \boldsymbol{\mu}_{k, t}^{s})^\top }{N_{k, t}^{s} + \sum_{i=1}^{N^{batch}} P_k(y = k \mid x_i)},
\end{aligned}
\end{equation}
where $N_{k, t-1}^s$ represents the effective sample size, which we define as the cumulative number of observed samples for class $k$ in branch $s$ up to step $t-1$, $s \in \{in, ex\}$ denotes the specific branch, and $x_i$ denotes the $i$-th image feature of $\mathcal{X}_{in}^{batch}$ or $\mathcal{X}_{ex}^{batch}$. To avoid introducing additional computations, we follow \cite{friedman1989regularized} by employing a shrinkage-based approach: $\hat{\mathbf{\Lambda}} = [(1 - \epsilon)\hat{\mathbf{\Sigma}} + \epsilon \mathbf{I}]^{-1}$, with $\epsilon = 10^{-4}$. 

\noindent\textit{(4) Test Time Classification with CLIP and GDA.}
\noindent The fused GDA logits of the inclusion and exclusion classifier are defined as: $f_{k}(x) = f_{k}^{in}(x) - \beta f_{k}^{ex}(x)$,
where $\beta \in [0,1]$ controls the weights of exclusion distributions. As the number of test samples increases and the estimated GDA distributions become more reliable, we integrate the logits from both the CLIP and GDA classifiers. Formally, the final fusion probability is defined as:
\begin{equation} \label{eq:fusion_classifier}
P_{t}(y=k\mid x)
=
\frac{
\exp\!\big(\cos(x,\mathbf{w}_{k})/\tau+\alpha_{t}\,(f_{k}^{in}(x) - \beta f_{k}^{ex}(x))\big)
}{
\sum_{j=1}^{K}
\exp\!\big(\cos(x,\mathbf{w}_{j})/\tau+\alpha_{t}\,(f_{j}^{in}(x) - \beta f_{j}^{ex}(x))\big)
},
\end{equation}
where $\tau$ is the temperature parameter and $\mathbf{w}_k$ is the $k$-th zero-shot text prototype. We adopt a dynamic weighting strategy where the parameter $\alpha_t$ balances the contribution of the GDA classifier. Specifically, $\alpha_t$ remains low when test samples are scarce and increases as more data is accumulated. This is defined as $\alpha_t = \min\left( \rho \cdot (B \cdot t), \alpha_{max} \right)$, where $\rho$ is a scaling factor, $B$ is the batch size, $t$ is the current iteration step, and $\alpha_{max}$ is a predefined upper bound (typically 1.0). This constraint prevents the GDA classifier from overly dominating the zero-shot prior, ensuring the model relies on the zero-shot classifier when the estimated test distribution remains unreliable.

\noindent\textbf{Negative Label Distribution Estimation.}
Since negative labels contain a large number of irrelevant negative labels (\cref{fig:neglabel_bias}), we leverage these pseudo samples to identify the negative labels that can accurately characterize the OOD distribution and effectively distinguish ID from OOD samples. Specifically, we estimate the negative labels distribution corresponding to negative images by computing the similarity scores, which are defined as follows:
\begin{equation} \label{eq:similarity_score}
\mathrm{Sim}(\mathcal{X}, y_i^{-}) = \frac{1}{|\mathcal{X}|} \sum_{x \in \mathcal{X}} \frac{\exp(x^\top \tilde{\mathbf{w}}_i / \tau)}{\sum_{j=1}^{K} \exp(x^\top \mathbf{w}_j / \tau) + \sum_{j=1}^{N} \exp(x^\top \tilde{\mathbf{w}}_j / \tau)}.
\end{equation}
An effective negative label will have a high similarity score on $\mathcal{X}_{neg}$ and a low similarity score on $\mathcal{X}_{pos}$.
After obtaining the similarity scores, we compute a contrastive score to evaluate the ability to distinguish negative images from positive images. This discriminative score is defined as follows:
\begin{equation} \label{eq:discriminative_score}
\Delta\mathrm{Sim}(y_i^{-})
=
\mathrm{Sim}(\mathcal{X}_{neg}, y_i^{-})
-
\mathrm{Sim}(\mathcal{X}_{pos}, y_i^{-}).
\end{equation}
Then we select the most discriminative $\hat{M}$ negative labels as follow:
\begin{equation} \label{eq:discriminative_neglabels_selection}
\mathcal{Y}^{*}
=
\operatorname{Top}
\left(
\{ \Delta\mathrm{Sim}(y_i^{-}) \}_{i=1}^{M},
\; \mathcal{Y}^{-},
\; \hat{M}
\right).
\end{equation}
We use the discriminative negative labels $\mathcal{Y}^{*}$ to calculate the confidence score: \\
\begin{equation} \label{eq:discriminative_neglabel_score}
S_{\text{nl}}(x) = \sum_{i=1}^{K} \left( \frac{\exp(x^\top \mathbf{w}_i / \tau)}{\sum_{j=1}^{K} \exp(x^\top \mathbf{w}_j / \tau) + \sum_{j=1}^{\hat{M}} \exp(x^\top \mathbf{w}^{*}_j / \tau)} \right).
\end{equation}
Here, $\mathbf{w}^{*}_j = f_{\text{txt}}(\rho(y^*_j)) \in \mathbb{R}^{D}$ denotes the text feature of the $y^{*}_j \in \mathcal{Y}^*$.\\

\noindent\textbf{Adaptive Threshold.}
The OOD detection process utilizes a scoring function $S(\cdot)$ to distinguish between ID and OOD inputs. Specifically, the OOD detector $G_\lambda(\cdot)$ is defined by a threshold $\lambda \in \mathbb{R}$, such that an input $x_i$ is classified as follows: $G_\lambda(x_i) = \text{Clean}$ if $S(x_i) \ge \lambda$, and $G_\lambda(x_i) = \text{Noise}$ otherwise. A test sample $x_i$ is classified as ID if $S(x_i)\ge\lambda$. A fixed threshold $\lambda$ generalizes poorly across diverse ID datasets in NTTA. To address this, \cite{li2023robustness} proposes an adaptive threshold that dynamically calibrates $\lambda$ by minimizing intra-class variance, leveraging the bimodal nature of OOD scores:
\begin{equation} \label{eq:adaptive_threshold}
\min_{\lambda} \frac{1}{Q_{id}} \sum_{i} \left[ S(x_i) - \frac{1}{Q_{id}} \sum_{j} \mathds{1}(S(x_j) > \lambda) S(x_j) \right]^2 + \frac{1}{Q_{ood}} \sum_{i} \left[ S(x_i) - \frac{1}{Q_{ood}} \sum_{j} \mathds{1}(S(x_j) \leq \lambda) S(x_j) \right]^2,
\end{equation}
\noindent where $Q_{id} = \sum_{i}^{Q} \mathds{1}(S(x_i) > \lambda)$ and $Q_{ood} = \sum_{i}^{Q} \mathds{1}(S(x_i) \le \lambda)$ are the lengths of the test time pseudo ID and OOD samples. 

\noindent The overall procedure of our DDE method is detailed in \cref{alg:dde}.

\begin{algorithm}[t]
\caption{Dual Distribution Estimation (DDE)}
\label{alg:dde}
\begin{algorithmic}[1]
\REQUIRE ID label space $\mathcal{Y}$, candidate negative labels $\mathcal{Y}^{-}$, batch data $\mathcal{X}^{batch}$.
\STATE Initialize dual Gaussian parameters $\mathcal{N}(\boldsymbol{\mu}_k^{\mathrm{in}}, \boldsymbol{\Sigma}_k^{\mathrm{in}})$ and $\mathcal{N}(\boldsymbol{\mu}_k^{\mathrm{ex}}, \boldsymbol{\Sigma}_k^{\mathrm{ex}})$;
\STATE Initialize two caches $\mathcal {C}_{pos}$ and $\mathcal {C}_{neg}$;

\FOR{each incoming batch $\mathcal{X}^{batch}$}
    \STATE \COMMENT{\textbf{Positive and Negative Images Selection}}
    \STATE Select positive/negative features and update caches $\mathcal{C}_{\text{pos}}$ and $\mathcal{C}_{\text{neg}}$ via~\cref{eq:pos_neg_images_cache};
    \STATE \COMMENT{\textbf{Positive Features Distribution Estimation}}
    \STATE Calculate class probability of GDA via~\cref{eq:gda_logits} and~\cref{eq:gda_classifier};
    \STATE Partition the Positive Images;
    \STATE Estimate inclusive and exclusive Gaussian Distribution Parameters 
    $(\boldsymbol{\mu}_k^{\mathrm{in}}, \boldsymbol{\Sigma}_k^{\mathrm{in}})$ and $(\boldsymbol{\mu}_k^{\mathrm{ex}}, \boldsymbol{\Sigma}_k^{\mathrm{ex}})$  via~\cref{eq:gauss_dis_estimation};
    \STATE Fuse classifier of CLIP and GDA to predict labels via~\cref{eq:fusion_classifier};
    \STATE \COMMENT{\textbf{Negative Labels Distribution Estimation}}
    \STATE Calculate similarity score $\mathrm{Sim}(\mathcal{X}, y_i^-)$ and discriminative score $\Delta\mathrm{Sim}(y_i^{-})$ via~\cref{eq:similarity_score} and~\cref{eq:discriminative_score};
    \STATE Select top-$\hat{M}$ discriminative negative labels via~\cref{eq:discriminative_neglabels_selection}; 
    \STATE Calculate the NegLabel OOD scores with discriminative negative labels via~\cref{eq:discriminative_neglabel_score};
    \STATE \COMMENT{\textbf{Adaptive Threshold}}
    \STATE Calculate the adaptive threshold and separate ID and OOD via~\cref{eq:adaptive_threshold}.
    
\ENDFOR
\STATE \textbf{Output:} Predicted labels and OOD scores.
\end{algorithmic}
\end{algorithm}

\begin{table}[t]
\centering
\caption{Zero-shot noisy TTA results for ImageNet-1k, ImageNet-S, ImageNet-A, ImageNet-V2, ImageNet-R as the ID datasets. \textbf{Bold} indicates the best performance.}
\label{tab:main_results}
\scriptsize
\resizebox{\columnwidth}{!}{
\begin{tabularx}{1.2\textwidth}{ll *{15}{>{\centering\arraybackslash}X}}
\toprule
\multirow{2}{*}{\textbf{ID}} & \multirow{2}{*}{\textbf{Method}} & \multicolumn{3}{c}{\textbf{iNaturalist}} & \multicolumn{3}{c}{\textbf{SUN}} & \multicolumn{3}{c}{\textbf{Texture}} & \multicolumn{3}{c}{\textbf{Places}} & \multicolumn{3}{c}{\textbf{Avg}} \\
\cmidrule(lr){3-5} \cmidrule(lr){6-8} \cmidrule(lr){9-11} \cmidrule(lr){12-14} \cmidrule(lr){15-17}
& & $Acc_S$ & $Acc_N$ & $Acc_H$ & $Acc_S$ & $Acc_N$ & $Acc_H$ & $Acc_S$ & $Acc_N$ & $Acc_H$ & $Acc_S$ & $Acc_N$ & $Acc_H$ & $Acc_S$ & $Acc_N$ & $Acc_H$ \\
\midrule
\multirow{6}{*}{ImageNet} 
& ZS-CLIP~\cite{radford2021learning} & 54.01 & 86.53 & 66.51 & 53.43 & 83.96 & 65.30 & 52.71 & 78.52 & 63.08 & 53.35 & 80.50 & 64.17 & 53.38 & 82.38 & 64.77 \\
& Tent~\cite{wang2020tent} & 48.56 & 35.74 & 41.18 & 55.44 & 75.54 & 63.95 & 54.94 & 70.93 & 61.92 & 55.76 & 73.98 & 63.59 & 53.67 & 64.05 & 57.66 \\
& TPT~\cite{shu2022test} & 52.58 & 88.93 & 66.09 & 51.91 & 86.09 & 64.77 & 51.11 & 80.01 & 62.38 & 51.80 & 82.89 & 63.76 & 51.85 & 84.48 & 64.25 \\
& DMN~\cite{zhang2024dual_cvpr} & 63.41 & 98.74 & 77.22 & 61.77 & 88.53 & 72.77 & 60.32 & 46.26 & 52.36 & 61.38 & 77.96 & 68.68 & 61.73 & 77.87 & 67.76 \\
& AdaNeg~\cite{zhang2024adaneg} & 62.00 & 99.21 & 76.31 & 61.14 & 92.95 & 73.76 & 61.40 & 86.65 & 71.87 & 61.36 & 82.23 & 70.28 & 61.48 & 90.26 & 73.06 \\
& OODD~\cite{yang2025oodd} & 59.40 & 99.50 & 74.39 & 60.01 & 90.28 & 72.10 & 59.99 & 77.84 & 67.76 & 59.92 & 79.51 & 68.34 & 59.83 & 86.78 & 70.65 \\
& AdaND~\cite{cao2025noisy} & 63.26 & 96.87 & 76.54 & 61.34 & 89.44 & 72.77 & 62.45 & 83.54 & 71.47 & 61.92 & 84.82 & 71.58 & 62.24 & 88.67 & 73.09 \\
\rowcolor{gray!15} & \textbf{DDE} & \textbf{67.72} & \textbf{99.45} & \textbf{80.57} & \textbf{65.57} & \textbf{96.25} & \textbf{78.00} & \textbf{63.60} & \textbf{94.04} & \textbf{75.88} & \textbf{65.31} & 82.02 & \textbf{72.71} & \textbf{65.55} & \textbf{92.94} & \textbf{76.79} \\
\midrule
\multirow{6}{*}{ImageNet-S} 
& ZS-CLIP~\cite{radford2021learning} & 34.17 & 83.46 & 48.49 & 33.46 & 81.20 & 47.39 & 32.61 & 75.57 & 45.56 & 33.40 & 77.10 & 46.61 & 33.41 & 79.33 & 47.01 \\
& Tent~\cite{wang2020tent} & 30.46 & 26.86 & 28.55 & 36.57 & 71.82 & 48.46 & 36.63 & 66.63 & 47.06 & 36.87 & 70.32 & 48.38 & 35.07 & 58.91 & 43.11 \\
& TPT~\cite{shu2022test} & 32.16 & 86.52 & 46.89 & 31.55 & 83.86 & 45.85 & 30.74 & 77.39 & 44.00 & 31.56 & 80.05 & 45.27 & 31.50 & 81.95 & 45.50 \\
& DMN~\cite{zhang2024dual_cvpr} & 42.69 & 98.36 & 59.54 & 41.54 & 87.94 & 56.42 & 40.80 & 75.12 & 52.88 & 41.32 & 78.53 & 54.14 & 41.59 & 84.99 & 55.75 \\
& AdaNeg\cite{zhang2024adaneg} & 39.90 & 99.82 & 57.01 & 39.90 & 96.23 & 56.41 & 39.21 & 89.01 & 54.44 & 39.90 & 89.11 & 55.12 & 39.73 & 93.54 & 55.75 \\
& OODD~\cite{yang2025oodd} & 42.41 & 98.31 & 59.26 & 42.81 & 69.54 & 53.00 & 32.83 & 89.75 & 48.08 & 32.67 & 89.17 & 47.82 & 37.68 & 86.69 & 52.04 \\
& AdaND~\cite{cao2025noisy} & 40.97 & 93.54 & 56.98 & 40.25 & 85.06 & 54.64 & 38.31 & 74.43 & 50.58 & 39.60 & 79.57 & 52.88 & 39.78 & 83.15 & 53.77 \\
\rowcolor{gray!15} & \textbf{DDE} & \textbf{43.23} & \textbf{99.47} & \textbf{60.27} & 41.42 & \textbf{96.55} & \textbf{57.97} & \textbf{43.61} & \textbf{89.15} & \textbf{58.57} & 40.42 & 83.59 & 54.49 & \textbf{42.17} & \textbf{92.19} & \textbf{57.83} \\
\midrule
\multirow{6}{*}{ImageNet-A} 
& ZS-CLIP~\cite{radford2021learning} & 34.73 & 80.69 & 48.56 & 34.20 & 78.83 & 47.70 & 33.97 & 76.60 & 47.07 & 33.96 & 75.11 & 46.77 & 34.22 & 77.81 & 47.53 \\
& Tent~\cite{wang2020tent} & 34.99 & 77.19 & 48.15 & 34.83 & 77.05 & 47.97 & 34.36 & 75.19 & 47.17 & 34.60 & 73.83 & 47.12 & 34.70 & 75.81 & 47.60 \\
& TPT~\cite{shu2022test} & 34.12 & 81.17 & 48.04 & 33.20 & 80.23 & 46.97 & 33.12 & 79.92 & 46.83 & 33.05 & 77.00 & 46.25 & 33.37 & 79.58 & 47.02 \\
& DMN~\cite{zhang2024dual_cvpr} & 42.59 & 97.17 & 59.22 & 41.90 & 84.57 & 56.04 & 40.62 & 58.74 & 48.03 & 40.95 & 74.45 & 52.84 & 41.52 & 78.73 & 54.03 \\
& AdaNeg\cite{zhang2024adaneg} & 42.55 & 98.58 & 59.45 & 42.11 & 86.94 & 56.74 & 42.15 & 92.30 & 57.88 & 41.46 & 75.23 & 53.46 & 42.07 & 88.26 & 56.88 \\
& OODD~\cite{yang2025oodd} & 41.50 & 99.22 & 58.52 & 36.23 & 92.73 & 52.10 & 36.94 & 91.22 & 52.58 & 36.18 & 72.01 & 48.16 & 37.71 & 88.80 & 52.84 \\
& AdaND~\cite{cao2025noisy} & 43.59 & 91.19 & 58.98 & 41.96 & 80.93 & 55.27 & 45.04 & 79.97 & 57.62 & 42.85 & 72.13 & 53.76 & 43.36 & 81.06 & 56.41 \\
\rowcolor{gray!15} & \textbf{DDE} & \textbf{48.85} & \textbf{99.20} & \textbf{65.46} & \textbf{47.37} & \textbf{94.38} & \textbf{63.08} & \textbf{51.14} & \textbf{90.21} & \textbf{65.28} & \textbf{46.70} & \textbf{76.83} & \textbf{58.09} & \textbf{48.51} & \textbf{90.16} & \textbf{62.98} \\
\midrule
\multirow{6}{*}{ImageNet-V2} 
& ZS-CLIP~\cite{radford2021learning} & 48.01 & 85.72 & 61.55 & 47.37 & 83.23 & 60.38 & 46.81 & 77.54 & 58.38 & 47.39 & 79.41 & 59.36 & 47.39 & 81.47 & 59.92 \\
& Tent~\cite{wang2020tent} & 47.94 & 76.98 & 59.08 & 48.28 & 80.50 & 60.36 & 47.56 & 74.47 & 58.05 & 48.34 & 77.37 & 59.50 & 48.03 & 77.33 & 59.25 \\
& TPT~\cite{shu2022test} & 46.63 & 88.37 & 61.05 & 46.12 & 85.58 & 59.94 & 45.21 & 79.14 & 57.55 & 46.02 & 81.95 & 58.94 & 46.00 & 83.76 & 59.37 \\
& DMN~\cite{zhang2024dual_cvpr} & 55.22 & 98.02 & 70.64 & 54.33 & 86.09 & 66.62 & 53.84 & 70.50 & 61.05 & 54.18 & 75.53 & 63.10 & 54.39 & 82.53 & 65.35 \\
& AdaNeg~\cite{zhang2024adaneg} & 52.36 & 99.43 & 66.27 & 50.73 & 95.53 & 66.27 & 51.21 & 88.88 & 64.98 & 50.75 & 86.70 & 64.02 & 51.26 & 92.64 & 65.97 \\
& OODD~\cite{yang2025oodd} & 53.01 & 99.45 & 69.16 & 53.60 & 90.48 & 67.32 & 55.53 & 77.80 & 63.42 & 53.57 & 79.52 & 64.02 & 53.43 & 86.81 & 65.98 \\
& AdaND~\cite{cao2025noisy} & 56.32 & 97.06 & 71.28 & 54.78 & 86.64 & 67.12 & 57.28 & 80.61 & 66.97 & 55.81 & 79.24 & 65.49 & 56.05 & 85.89 & 67.72 \\
\rowcolor{gray!15} & \textbf{DDE} & \textbf{59.84} & \textbf{99.52} & \textbf{74.74} & \textbf{56.64} & \textbf{97.51} & \textbf{71.66} & \textbf{55.41} & \textbf{94.77} & \textbf{69.93} & \textbf{56.72} & \textbf{87.52} & \textbf{68.83} & \textbf{57.15} & \textbf{94.83} & \textbf{71.29} \\
\midrule
\multirow{6}{*}{ImageNet-R} 
& ZS-CLIP~\cite{radford2021learning} & 61.99 & 94.39 & 74.83 & 61.82 & 88.95 & 72.94 & 60.91 & 77.05 & 68.04 & 61.68 & 84.86 & 71.44 & 61.60 & 86.31 & 71.81 \\
& Tent~\cite{wang2020tent} & 65.22 & 91.45 & 76.14 & 65.06 & 85.61 & 73.93 & 63.33 & 69.99 & 66.49 & 64.93 & 82.38 & 72.62 & 64.64 & 82.36 & 72.30 \\
& TPT~\cite{shu2022test} & 60.95 & 94.80 & 74.20 & 60.85 & 89.98 & 72.60 & 59.98 & 77.79 & 67.73 & 60.67 & 85.79 & 71.08 & 60.61 & 87.09 & 71.40 \\
& DMN~\cite{zhang2024dual_cvpr} & 69.44 & 98.43 & 81.43 & 69.23 & 92.78 & 79.29 & 68.94 & 88.12 & 77.36 & 69.00 & 83.93 & 75.73 & 69.15 & 90.81 & 78.46 \\
& AdaNeg~\cite{zhang2024adaneg} & 71.55 & 99.51 & 83.24 & 71.36 & 96.65 & 82.10 & 70.54 & 87.30 & 78.03 & 71.28 & 85.01 & 77.54 & 71.18 & 92.12 & 80.23 \\
& OODD~\cite{yang2025oodd} & 68.52 & 99.85 & 81.27 & 69.76 & 95.41 & 80.59 & 69.58 & 91.42 & 79.02 & 69.62 & 85.20 & 76.63 & 69.37 & 92.97 & 79.38 \\
& AdaND~\cite{cao2025noisy} & 72.21 & 99.59 & 83.72 & 71.02 & 95.94 & 81.62 & 70.44 & 81.43 & 75.54 & 70.85 & 92.14 & 80.10 & 71.13 & 92.28 & 80.25 \\
\rowcolor{gray!15} & \textbf{DDE} & \textbf{75.84} & \textbf{98.88} & \textbf{85.84} & \textbf{76.05} & \textbf{97.26} & \textbf{85.35} & 69.06 & \textbf{93.65} & \textbf{79.50} & \textbf{75.61} & 90.18 & \textbf{82.25} & \textbf{74.14} & \textbf{94.99} & \textbf{83.24} \\
\bottomrule
\end{tabularx}
}
\end{table}

\section{Experiments}
\subsection{Experimental Setup} 
\noindent\textbf{Datasets and Benchmarks.}
We evaluate our method under the NTTA setting, using ImageNet-1K~\cite{deng2009imagenet} as the primary ID dataset along with its distribution-shifted variants, namely ImageNet-S~\cite{wang2019learning}, ImageNet-A~\cite{hendrycks2021natural}, ImageNet-V2~\cite{recht2019imagenet}, and ImageNet-R~\cite{hendrycks2021many}. To assess generalization beyond ImageNet-style categories, we also incorporate fine-grained ID datasets, including CUB-200-2011~\cite{wah2011caltech}, Stanford Cars~\cite{krause20133d}, Food-101~\cite{bossard2014food}, and Oxford-IIIT Pet~\cite{parkhi2012cats}. For OOD evaluation, we utilize iNaturalist~\cite{van2018inaturalist}, SUN~\cite{xiao2010sun}, Texture~\cite{cimpoi2014describing}, and Places~\cite{zhou2017places} . Following the NTTA protocol, the test stream is constructed by sequentially mixing ID and noisy samples, which include both OOD types. All experiments are conducted in a zero-shot and source-free manner.

\noindent\textbf{Implementation Details.} We use the visual encoder of ViT-B/16 pretrained by CLIP~\cite{radford2021learning} for all experiments, while DDE performs fully training-free online adaptation. For the Positive and Negative Image Selection stage, we follow the configuration of AdaNeg~\cite{zhang2024adaneg}, setting $\lambda_{pos}=0.75$ and $\lambda_{neg}=0.25$ for all the experiments. The detailed setting analysis can be found in the Supplementary Material. The queue length is maintained at $Q=1000$. For the PFDE stage, we set the exclusion GDA weight $\beta=0.5$ and the scaling factor $\rho=0.005$. In the NLDE stage, we employ $\hat{M}=500$ for the ImageNet benchmark~\cite{deng2009imagenet} and $\hat{M}=100$ for fine-grained dataset.

\noindent\textbf{Evaluation Metrics.} In the noisy TTA setting, we follow OWTTT~\cite{li2023robustness} to report ID accuracy ($Acc_S$), noisy detection accuracy ($Acc_N$), and their harmonic mean ($Acc_H$). The definition details are available in the Supplementary Material.

\begin{table*}[t]
\centering
\caption{Zero-shot noisy TTA results for CUB-200-2011, STANFORD-CARS, Food-101, and Oxford-IIIT Pet as the ID datasets. \textbf{Bold} indicates the best performance.}
\label{tab:fine_grained_tta}
\scriptsize
\resizebox{\textwidth}{!}{
\begin{tabularx}{1.2\textwidth}{ll *{15}{>{\centering\arraybackslash}X}}
\toprule
\multirow{2}{*}{\textbf{Dataset}} & \multirow{2}{*}{\textbf{Method}} & \multicolumn{3}{c}{\textbf{iNaturalist}} & \multicolumn{3}{c}{\textbf{SUN}} & \multicolumn{3}{c}{\textbf{Texture}} & \multicolumn{3}{c}{\textbf{Places}} & \multicolumn{3}{c}{\textbf{Avg}} \\
\cmidrule(lr){3-5} \cmidrule(lr){6-8} \cmidrule(lr){9-11} \cmidrule(lr){12-14} \cmidrule(lr){15-17}
& & $Acc_S$ & $Acc_N$ & $Acc_H$ & $Acc_S$ & $Acc_N$ & $Acc_H$ & $Acc_S$ & $Acc_N$ & $Acc_H$ & $Acc_S$ & $Acc_N$ & $Acc_H$ & $Acc_S$ & $Acc_N$ & $Acc_H$ \\
\midrule
\multirow{6}{*}{\makecell[l]{CUB-200-\\2011}} 
& ZS-CLIP~\cite{radford2021learning} & 38.13 & 88.06 & 53.22 & 38.10 & 87.86 & 53.15 & 37.56 & 79.11 & 50.94 & 38.00 & 87.81 & 53.04 & 37.95 & 85.71 & 52.59 \\
& Tent~\cite{wang2020tent}& 37.02 & 46.95 & 41.40 & 38.61 & 55.55 & 45.56 & 34.98 & 41.77 & 38.07 & 40.41 & 74.83 & 52.48 & 37.75 & 54.78 & 44.38 \\
& TPT~\cite{shu2022test} & 37.41 & 89.57 & 52.78 & 37.49 & 89.67 & 52.87 & 36.88 & 81.67 & 50.81 & 37.44 & 89.45 & 52.79 & 37.30 & 87.59 & 52.31 \\
& DMN~\cite{zhang2024dual_cvpr} & 55.96 & 98.66 & 71.41 & 56.02 & 98.72 & 71.48 & 56.19 & 99.36 & 71.78 & 56.11 & 96.52 & 70.96 & 56.07 & 98.32 & 71.41 \\
& AdaNeg~\cite{zhang2024adaneg} & 56.82 & 99.65 & 72.38 & 56.79 & 99.14 & 72.21 & 56.96 & 93.37 & 70.75 & 56.92 & 99.31 & 72.37 & 56.87 & 97.87 & 71.93 \\
& OODD~\cite{yang2025oodd} & 56.06 & 99.73 & 71.77 & 56.02 & 99.84 & 71.77 & 56.06 & 99.91 & 71.82 & 56.06 & 98.27 & 71.39 & 56.05 & 99.44 & 71.69 \\
& AdaND~\cite{cao2025noisy} & 52.34 & 96.40 & 67.84 & 52.41 & 93.91 & 67.27 & 51.82 & 81.24 & 63.28 & 51.82 & 91.51 & 66.17 & 52.10 & 90.77 & 66.14 \\
\rowcolor{gray!15} & \textbf{DDE} 
& \textbf{58.56} & \textbf{99.58} & \textbf{73.77} 
& \textbf{58.74} & \textbf{99.55} & \textbf{73.91} 
& \textbf{59.89} & \textbf{99.80} & \textbf{74.88} 
& \textbf{58.47} & \textbf{96.28} & \textbf{72.71} 
& \textbf{56.92} & \textbf{98.80} & \textbf{73.82} \\

\midrule
\multirow{6}{*}{\makecell[l]{Stanford\\-CARS}}
& ZS-CLIP~\cite{radford2021learning} & 50.18 & 96.62 & 66.05 & 53.48 & 98.81 & 69.40 & 53.59 & 99.05 & 69.55 & 53.36 & 98.05 & 69.11 & 52.65 & 98.13 & 68.53 \\
& Tent~\cite{wang2020tent} & 44.12 & 52.33 & 47.88 & 54.27 & 94.51 & 68.95 & 54.60 & 97.37 & 69.97 & 54.33 & 96.65 & 69.56 & 51.83 & 85.22 & 64.09 \\
& TPT~\cite{shu2022test} & 49.24 & 96.97 & 65.31 & 52.40 & 98.83 & 68.49 & 52.75 & 99.27 & 68.89 & 52.42 & 98.39 & 68.40 & 51.70 & 98.36 & 67.77 \\
& DMN~\cite{zhang2024dual_cvpr} & 64.33 & 99.99 & 78.29 & 64.35 & 99.63 & 78.19 & 64.35 & 99.70 & 78.21 & 64.53 & 98.29 & 77.91 & 64.39 & 99.40 & 78.15 \\
& AdaNeg~\cite{zhang2024adaneg} & 63.66 & 99.98 & 77.80 & 63.51 & 99.89 & 77.65 & 63.71 & 99.87 & 77.79 & 63.54 & 99.66 & 77.60 & 63.61 & 99.85 & 77.71 \\
& OODD~\cite{yang2025oodd} & 63.55 & 99.99 & 77.71 & 63.55 & 99.88 & 77.68 & 63.56 & 99.95 & 77.71 & 63.55 & 99.04 & 77.42 & 63.55 & 99.72 & 77.63 \\
& AdaND~\cite{cao2025noisy} & 62.80 & 99.79 & 77.09 & 62.73 & 99.82 & 77.04 & 62.91 & 99.75 & 77.16 & 62.76 & 99.29 & 76.91 & 62.80 & 99.66 & 77.05 \\
\rowcolor{gray!15} & \textbf{DDE} 
& \textbf{65.51} & \textbf{99.99} & \textbf{79.16} 
& \textbf{65.30} & \textbf{99.89} & \textbf{78.98} 
& \textbf{66.35} & \textbf{99.93} & \textbf{79.75} 
& \textbf{65.43} & 98.66 & \textbf{78.68} 
& \textbf{65.65} & 99.62 & \textbf{79.14} \\
\midrule
\multirow{6}{*}{Food-101} 
& ZS-CLIP~\cite{radford2021learning} & 80.60 & 94.76 & 87.11 & 80.75 & 96.08 & 87.75 & 80.51 & 93.12 & 86.36 & 80.62 & 94.62 & 87.06 & 80.62 & 94.65 & 87.07 \\
& Tent~\cite{wang2020tent} & 75.83 & 25.09 & 37.70 & 82.86 & 85.10 & 83.97 & 82.54 & 87.03 & 84.73 & 82.26 & 80.13 & 81.18 & 80.87 & 69.34 & 71.90 \\
& TPT~\cite{shu2022test}& 79.70 & 94.93 & 86.65 & 79.92 & 96.19 & 87.30 & 79.70 & 93.86 & 86.20 & 79.76 & 95.14 & 86.77 & 79.77 & 95.03 & 86.73 \\
& DMN~\cite{zhang2024dual_cvpr} & 84.17 & 99.92 & 91.37 & 84.13 & 99.87 & 91.33 & 84.14 & 95.71 & 89.55 & 84.21 & 99.49 & 91.21 & 84.16 & 98.75 & 90.87 \\
& AdaNeg~\cite{zhang2024adaneg} & 84.17 & 99.74 & 91.29 & 84.16 & 98.86 & 90.92 & 84.23 & 96.82 & 90.09 & 84.16 & 93.65 & 88.65 & 84.18 & 97.27 & 90.24 \\
& OODD~\cite{yang2025oodd} & 83.63 & 99.99 & 91.09 & 83.77 & 99.99 & 91.17 & 83.75 & 97.22 & 89.98 & 83.76 & 99.91 & 91.13 & 83.73 & 99.28 & 90.84 \\
& AdaND~\cite{cao2025noisy} & 86.50 & 99.87 & 92.71 & 86.40 & 99.64 & 92.55 & 86.44 & 96.51 & 91.20 & 86.42 & 99.40 & 92.46 & 86.44 & 98.85 & 92.23 \\
\rowcolor{gray!15} & \textbf{DDE} 
& \textbf{88.35} & \textbf{99.93} & \textbf{93.78} 
& \textbf{88.23} & \textbf{99.78} & \textbf{93.65} 
& \textbf{88.30} & \textbf{97.94} & \textbf{92.87} 
& \textbf{88.24} & \textbf{99.65} & \textbf{93.61} 
& \textbf{88.28} & \textbf{99.33} & \textbf{93.48} \\
\midrule
\multirow{6}{*}{\makecell[l]{Oxford-IIIT\\ Pet}} 
& ZS-CLIP~\cite{radford2021learning} & 78.58 & 88.30 & 83.16 & 79.75 & 87.30 & 83.35 & 80.20 & 91.16 & 85.33 & 79.59 & 84.17 & 81.82 & 79.53 & 87.73 & 83.41 \\
& Tent~\cite{wang2020tent} & 80.07 & 78.09 & 79.07 & 81.19 & 68.30 & 74.19 & 81.48 & 74.72 & 77.95 & 80.64 & 62.51 & 70.43 & 80.84 & 70.91 & 75.41 \\
& TPT~\cite{shu2022test} & 77.56 & 89.71 & 83.19 & 78.87 & 89.82 & 83.99 & 79.17 & 92.26 & 85.22 & 78.62 & 87.32 & 82.74 & 78.56 & 89.78 & 83.78 \\
& DMN~\cite{zhang2024dual_cvpr} & 87.98 & 97.62 & 92.55 & 87.90 & 97.55 & 92.47 & 88.17 & 96.77 & 92.27 & 88.12 & 96.91 & 92.31 & 88.04 & 97.21 & 92.40 \\
& AdaNeg~\cite{zhang2024adaneg} & 88.06 & 99.49 & 93.43 & 87.98 & 91.72 & 89.81 & 88.25 & 94.56 & 91.30 & 88.23 & 91.29 & 89.73 & 88.13 & 94.26 & 91.07 \\
& OODD~\cite{yang2025oodd} & 85.04 & 99.99 & 91.91 & 85.58 & 99.96 & 92.21 & 85.69 & 99.82 & 92.22 & 85.58 & 99.74 & 92.12 & 85.47 & 99.88 & 92.12 \\
& AdaND~\cite{cao2025noisy} & 85.81 & 98.78 & 91.84 & 85.82 & 98.19 & 91.59 & 85.86 & 98.68 & 91.82 & 85.88 & 96.58 & 90.92 & 85.84 & 98.06 & 91.54 \\
\rowcolor{gray!15} & \textbf{DDE} & \textbf{88.28} & \textbf{99.97} & \textbf{93.76} & \textbf{88.72} & 99.14 & \textbf{93.64} & \textbf{88.74} & \textbf{99.63} & \textbf{93.87} & \textbf{88.58} & \textbf{96.88} & \textbf{92.54} & \textbf{88.58} & 98.90 & \textbf{93.45} \\
\bottomrule
\end{tabularx}
}
\vspace{-0.2cm}
\end{table*}

\subsection{Main Results}

\noindent\textbf{Zero-Shot Noisy TTA on ImageNet and its Variants.}
\cref{tab:main_results} presents the results for ImageNet and its distribution-shifted variants. DDE consistently achieves the highest harmonic mean accuracy ($Acc_H$) across all five ID datasets. Specifically, on ImageNet, DDE improves the average $Acc_H$ to 76.79\%, outperforming the strongest baseline, AdaND (73.09\%), by 3.70\%. DDE simultaneously improves ID accuracy and noisy detection performance, reaching 67.73\% $Acc_S$ and 99.42\% $Acc_N$ on iNaturalist. Similar improvements are observed on ImageNet-A (+6.57\% over AdaND), ImageNet-V2 (+3.57\%), and ImageNet-R (+3.01\%). These findings suggest that dual distribution modeling effectively balances classification and noise detection under distribution shifts.

\noindent\textbf{Zero-Shot Noisy TTA across Various ID Datasets.}
\cref{tab:fine_grained_tta} evaluates generalization to fine-grained ID datasets. DDE consistently delivers the best overall performance across CUB-200-2011, Stanford Cars, Food-101, and Oxford-IIIT Pet. For example, on Food-101, DDE achieves an average $Acc_H$ of 93.48\%, outperforming AdaND (92.23\%). On Stanford Cars, DDE reaches 79.14\% average $Acc_H$, improving over AdaND (77.05\%). The gains are particularly significant on the challenging fine-grained datasets such as CUB, where accurately modeling intra-class variance is essential. These results demonstrate that our positive feature distribution estimation generalizes beyond ImageNet-style categories and remains robust in fine-grained recognition settings.

\begin{table}[t]
\centering
\caption{Zero-shot OOD detection results for ImageNet as the ID dataset. \textbf{Bold} indicates the best performance.}
\label{tab:ood_detection}
\scriptsize
\resizebox{\columnwidth}{!}{
\begin{tabularx}{1.1\textwidth}{l *{10}{>{\centering\arraybackslash}X}}
\toprule
\multirow{2}{*}{\textbf{Method}} & \multicolumn{2}{c}{\textbf{iNaturalist}} & \multicolumn{2}{c}{\textbf{SUN}} & \multicolumn{2}{c}{\textbf{Texture}} & \multicolumn{2}{c}{\textbf{Places}} & \multicolumn{2}{c}{\textbf{Avg}} \\
\cmidrule(lr){2-3} \cmidrule(lr){4-5} \cmidrule(lr){6-7} \cmidrule(lr){8-9} \cmidrule(lr){10-11}
& AUROC$\uparrow$ & FPR95$\downarrow$ & AUROC$\uparrow$ & FPR95$\downarrow$ & AUROC$\uparrow$ & FPR95$\downarrow$ & AUROC$\uparrow$ & FPR95$\downarrow$ & AUROC$\uparrow$ & FPR95$\downarrow$ \\
\midrule
Max-Logit & 89.31 & 61.66 & 87.43 & 64.39 & 71.68 & 86.61 & 85.95 & 63.67 & 83.59 & 69.08 \\
Energy    & 85.09 & 81.08 & 84.24 & 79.02 & 65.56 & 93.65 & 83.38 & 75.08 & 79.57 & 82.21 \\
MCM       & 94.61 & 30.91 & 92.57 & 37.59 & 86.11 & 57.77 & 89.77 & 44.69 & 90.77 & 42.74 \\
NegLabel  & 99.49 & 1.91  & 95.49 & 20.53 & 90.22 & 43.56 & 91.64 & 35.59 & 94.21 & 25.40 \\
NegRefine  & 99.57 & 1.51  & 94.63 & 22.93 & 94.68 & 21.15 & 90.41 & 39.10 & 94.82 & 21.16 \\
OODD     & 99.74 & 0.51  & 96.84 & 16.30 & 93.19 & 34.09 & 93.77 & 28.68 & 95.89 & 19.90 \\
AdaNeg     & 98.71 & 0.59  & 97.44 & 9.50 & 94.55 & 34.34 & 94.93 & 31.27 & 96.66 & 18.92 \\
AdaND     & 98.91 & 4.19  & 95.86 & 17.08 & 93.01 & 21.76 & 94.55 & 20.95 & 95.58 & 16.00 \\
\rowcolor{gray!15} \textbf{DDE} & \textbf{99.83} & \textbf{0.47} & \textbf{98.97} & \textbf{3.45} & \textbf{95.98} & \textbf{20.85} & \textbf{96.78} & \textbf{14.42} & \textbf{97.89} & \textbf{9.80} \\
\bottomrule
\end{tabularx}
}
\end{table}

\noindent\textbf{Traditional OOD Detection.}
\Cref{tab:ood_detection} reports zero-shot OOD detection results on ImageNet. DDE achieves the best overall performance, reaching 97.89\% AUROC and reducing the average FPR95 to 9.80\%, substantially outperforming AdaND (95.58\% AUROC, 16.00\% FPR95). These results demonstrate the effectiveness of dynamically estimating discriminative negative semantic distributions for OOD separation.

\subsection{Analyses and Discussions}
\noindent\textbf{Ablation Study.} \cref{tab:ablation} illustrates the contribution of each module in DDE. Specifically, replacing PFDE (in) with PFDE (in+ex) improves the harmonic mean from $72.77\%$ to $73.52\%$ in the noisy stream, confirming the benefit of modeling exclusion distributions. Furthermore, NLDE significantly boosts noisy sample detection, increasing $\text{Acc}_N$ from $83.96\%$ to $96.25\%$. The full DDE (incorporating Ada $\lambda$) achieves the best balance, reaching the highest harmonic mean of \textbf{78.00\%}. This demonstrates that the three modules work synergistically to handle noise effectively.


\begin{table}[t]
\centering
\caption{\textbf{Ablation of DDE components on ImageNet} under noisy and clean streams. For PFDE, ``(in)'' denotes using only the inclusion distribution, while ``(in+ex)'' uses both inclusion and exclusion distributions.}
\label{tab:ablation}
\scriptsize
\begin{tabular}{cccc|ccc|ccc}
\toprule
\multicolumn{4}{c|}{\textbf{Components}} & \multicolumn{3}{c|}{\textbf{Clean Stream}} & \multicolumn{3}{c}{\textbf{Noisy Stream}} \\
\cmidrule(lr){5-7} \cmidrule(lr){8-10}
\begin{tabular}[c]{@{}c@{}}\textbf{PFDE} \\ \textbf{(in)}\end{tabular} & \begin{tabular}[c]{@{}c@{}}\textbf{PFDE} \\ \textbf{(in+ex)}\end{tabular} & \textbf{NLDE} & \textbf{Ada$\lambda$} & \textbf{$\text{Acc}_S \uparrow$} & \textbf{$\text{Acc}_N \uparrow$} & \textbf{$\text{Acc}_H \uparrow$} & \textbf{$\text{Acc}_S \uparrow$} & \textbf{$\text{Acc}_N \uparrow$} & \textbf{$\text{Acc}_H \uparrow$} \\ \midrule
& & & & 66.62 & -- & -- & 53.43 & 83.96 & 65.30 \\
\checkmark & & & & 69.62 & -- & -- & 61.87 & 88.34 & 72.77 \\
& \checkmark  & & & 70.34 & -- & -- & 62.69 & 88.90 & 73.52 \\
& & \checkmark & & 66.83 & -- & -- & 55.72 & 95.99 & 70.51 \\
& \checkmark & \checkmark & & 70.81 & -- & -- & 64.14 & 95.76 & 76.82 \\
& \checkmark & \checkmark & \checkmark & 71.14 & -- & -- & \textbf{65.57} & \textbf{96.25} & \textbf{78.00} \\ \bottomrule
\end{tabular}
\end{table}

\begin{figure*}[t]
    \centering
    \begin{subfigure}{0.24\textwidth}
        \centering
        \includegraphics[width=\linewidth]{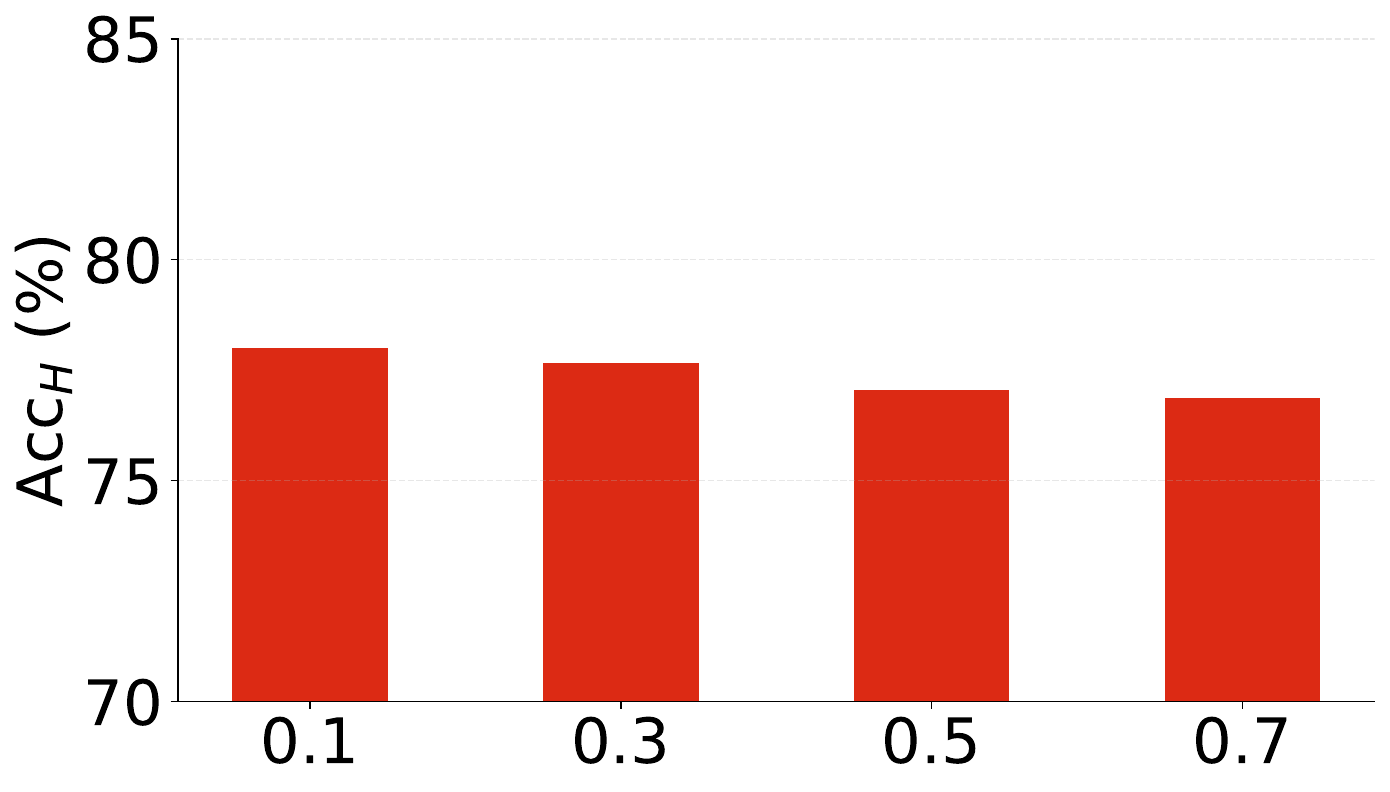}
        \caption{Cache Noise Rate}
        \label{fig:noise_rate}
    \end{subfigure}
    \hfill
    \begin{subfigure}{0.24\textwidth}
        \centering
        \includegraphics[width=\linewidth]{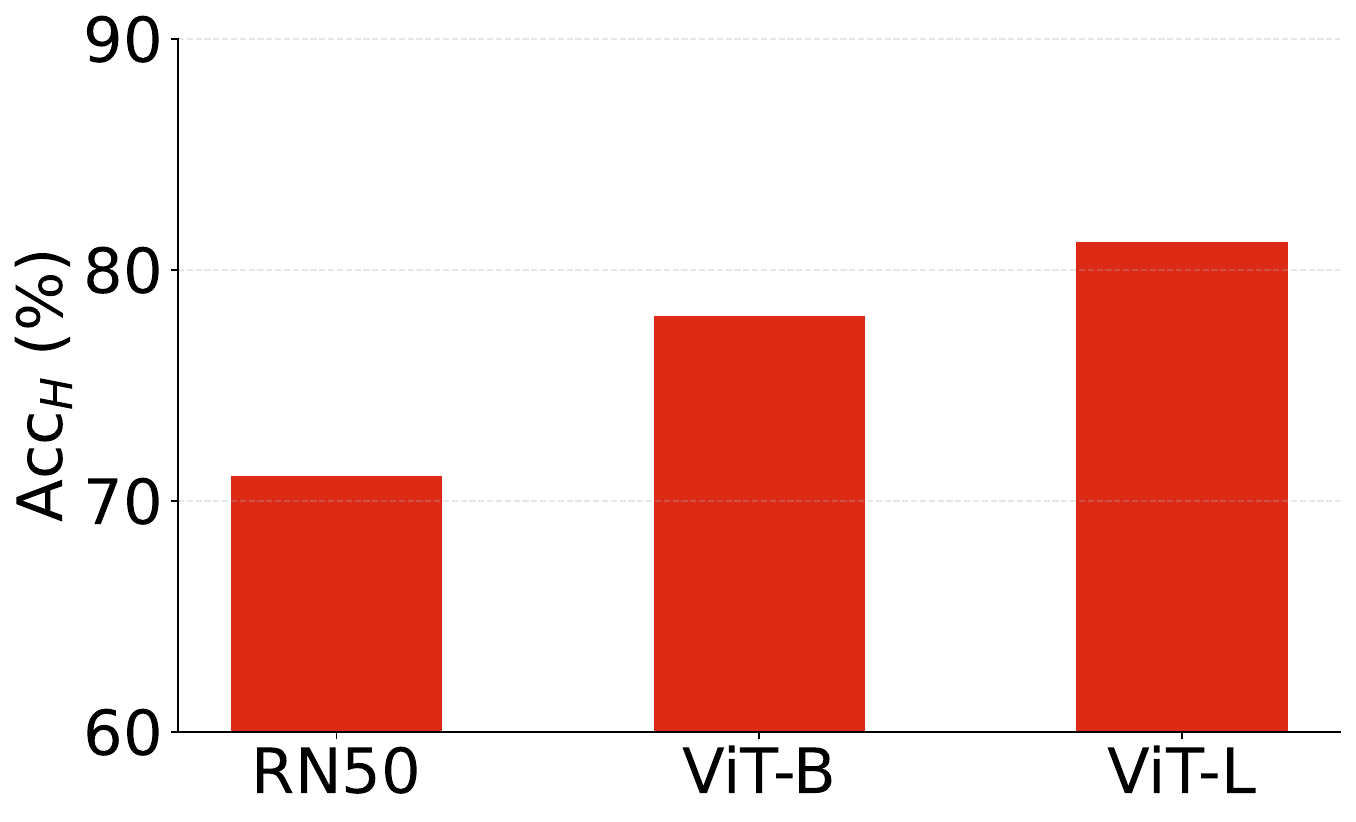}
        \caption{Backbones}
        \label{fig:backbones}
    \end{subfigure}
    \hfill
    \begin{subfigure}{0.24\textwidth}
        \centering
        \includegraphics[width=\linewidth]{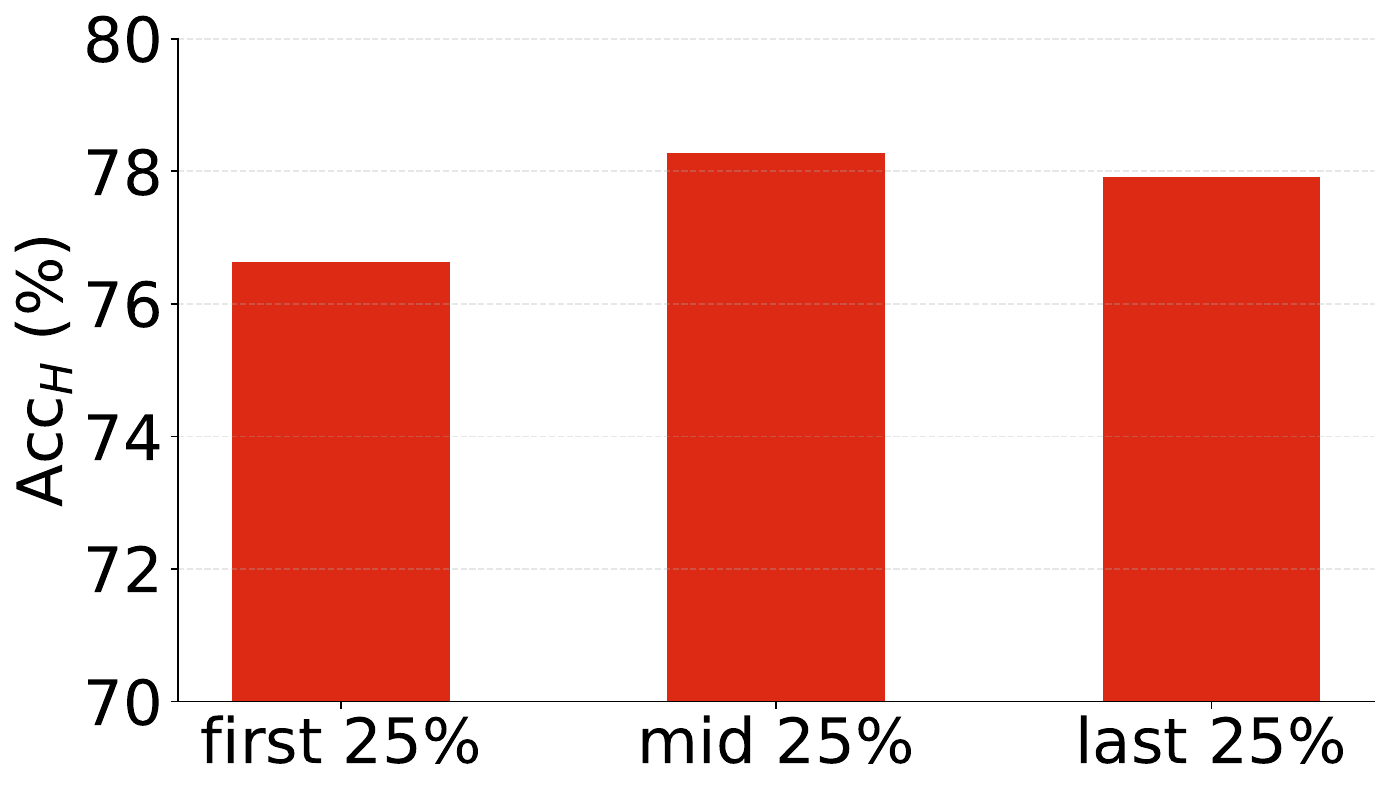}
        \caption{Stability}
        \label{fig:stability}
    \end{subfigure}
    \hfill
    \begin{subfigure}{0.24\textwidth}
        \centering
        \includegraphics[width=\linewidth]{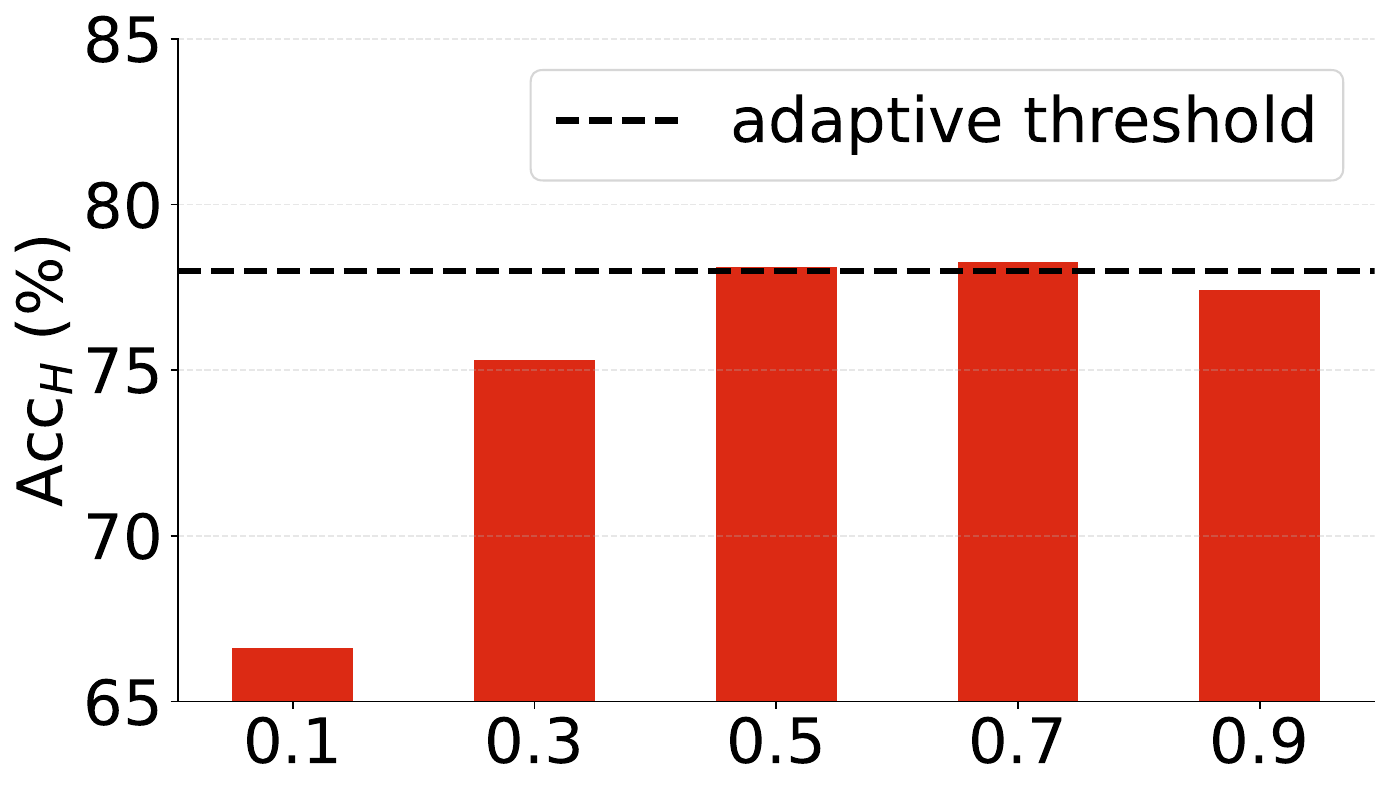}
        \caption{Fixed threshold}
        \label{fig:fixed_threshold}
    \end{subfigure}

    \begin{subfigure}{0.24\textwidth}
        \centering
        \includegraphics[width=\linewidth]{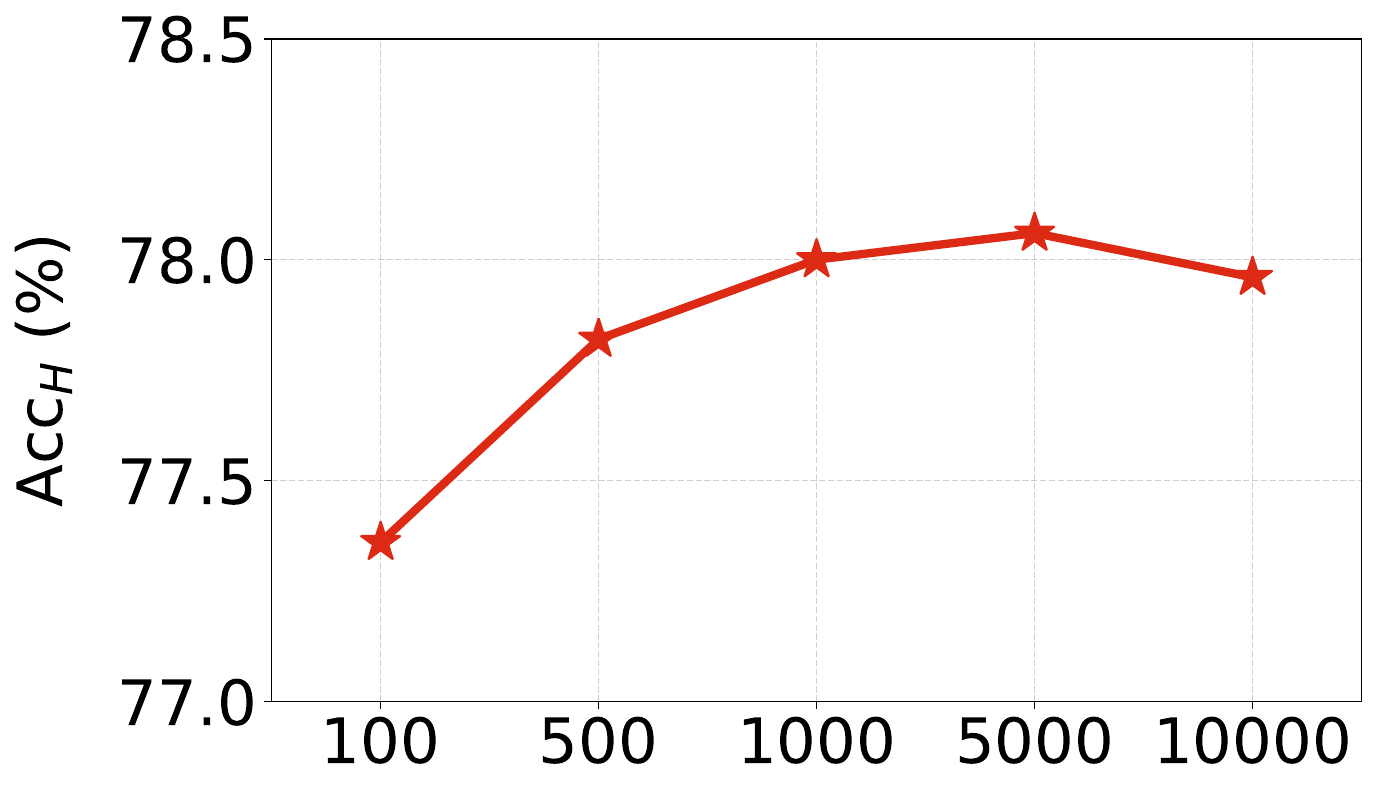}
        \caption{$Q$}
        \label{fig:Q}
    \end{subfigure}
    \hfill
    \begin{subfigure}{0.24\textwidth}
        \centering
        \includegraphics[width=\linewidth]{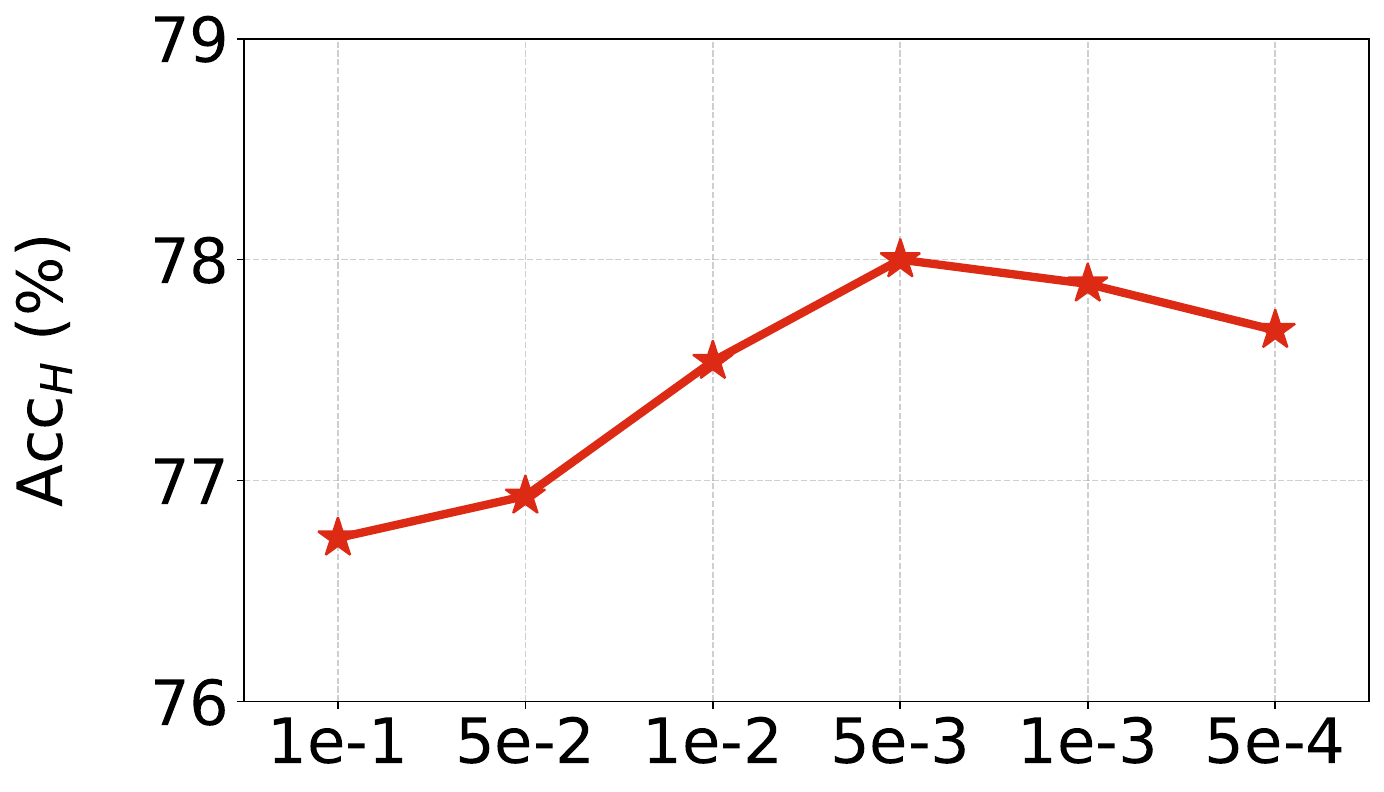}
        \caption{$\rho$}
        \label{fig:rho}
    \end{subfigure}
    \hfill
    \begin{subfigure}{0.24\textwidth}
        \centering
        \includegraphics[width=\linewidth]{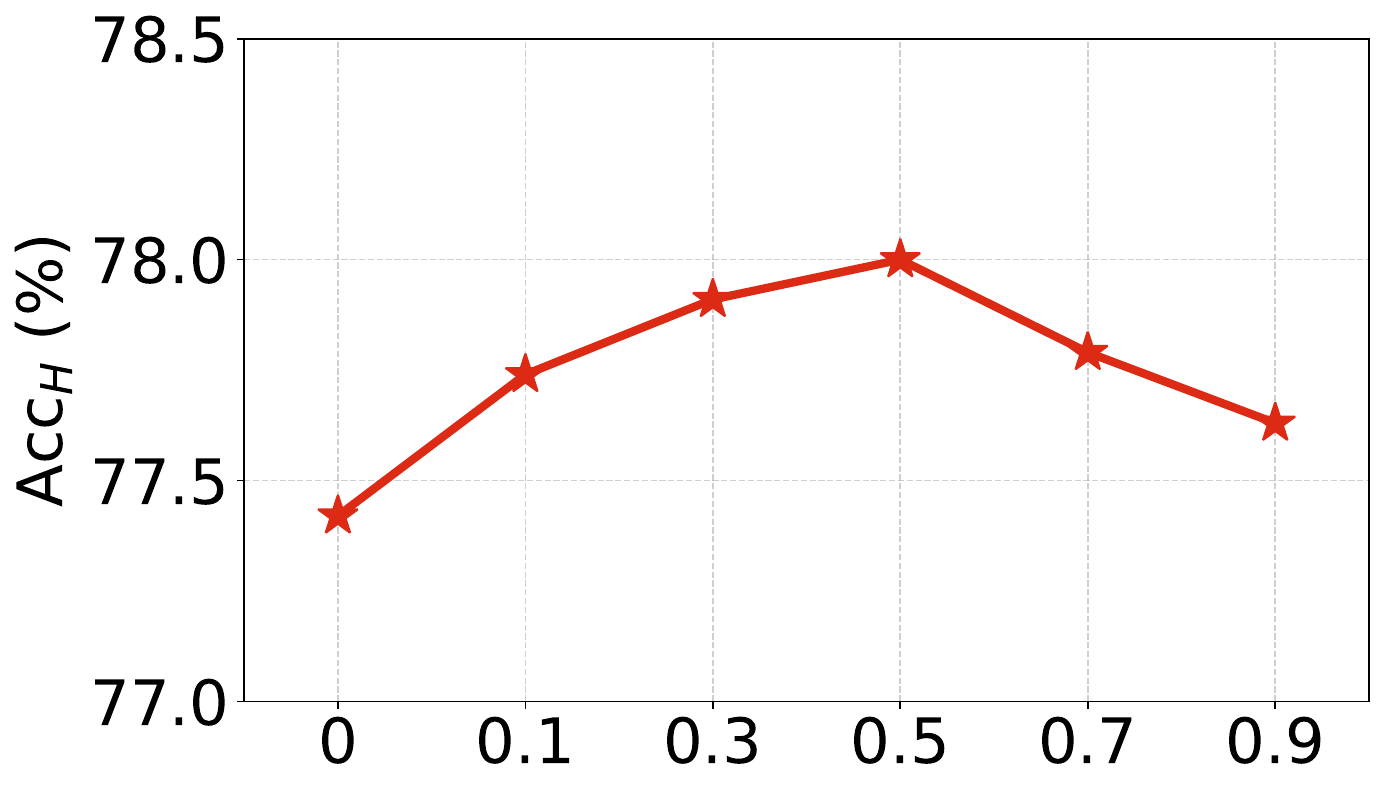}
        \caption{$\beta$}
        \label{fig:beta}
    \end{subfigure}
    \hfill
    \begin{subfigure}{0.24\textwidth}
        \centering
        \includegraphics[width=\linewidth]{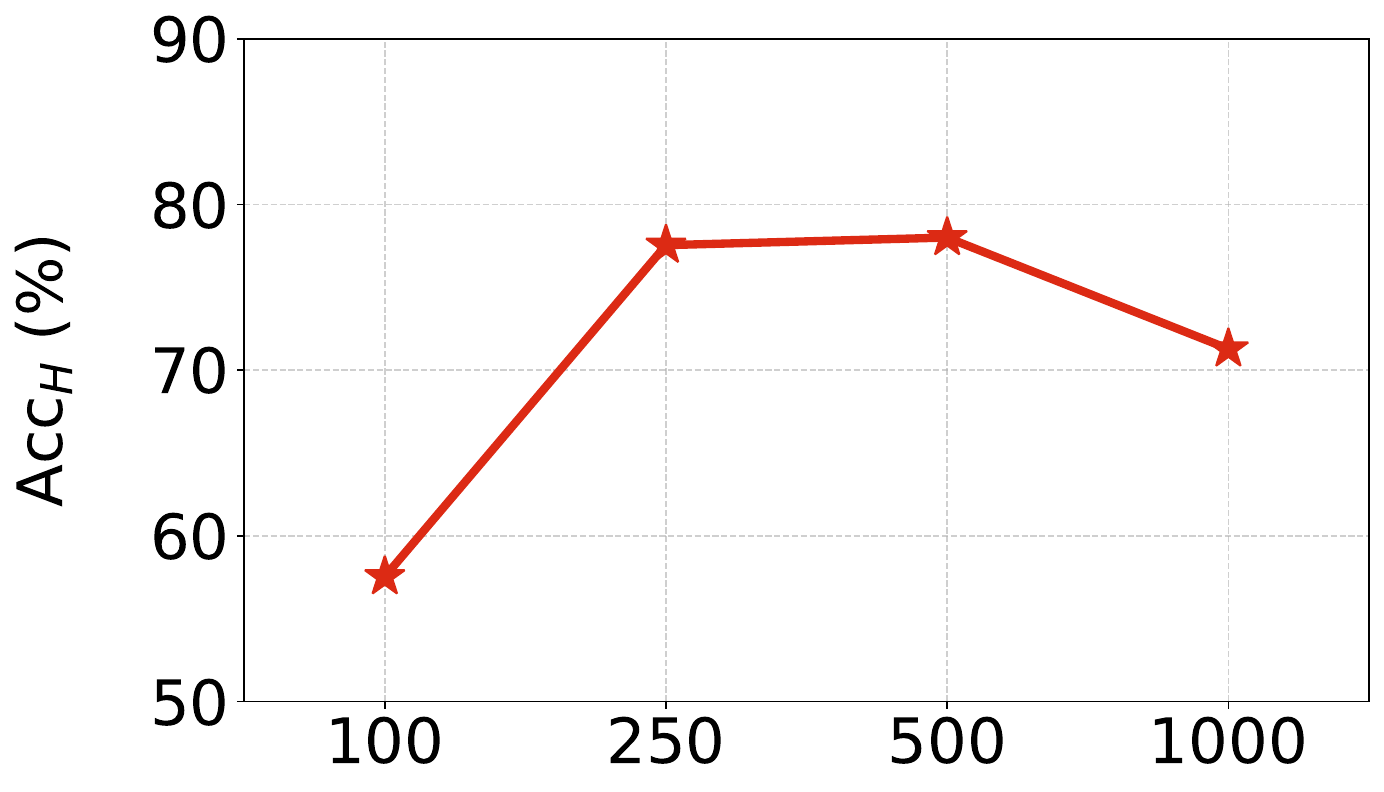}
        \caption{$\hat{M}$}
        \label{fig:M}
    \end{subfigure}
    \caption{\textbf{Comprehensive analysis of model components and hyperparameters.} We evaluate the harmonic mean ($\text{Acc}_H$) across various settings: (a) noise ratios of positive and negative images, (b) backbones, (c) stability, and (d) fixed threshold. The bottom row shows sensitivity analysis for (e) cache size of positive and negative images, (f) scaling factor $\rho$, (g) exclusion GDA model weight $\beta$, and (h) number of discriminative negative labels $\hat{M}$. All experiments are conducted on ImageNet.}
    \label{fig:comprehensive_analysis}
\end{figure*}

\noindent\textbf{Robustness to Noise.}
As shown in \cref{fig:noise_rate}, the model exhibits remarkable robustness, with $\text{Acc}_H$ decreasing only marginally from 78.00\% to 76.87\% as the noise rate increases from 0.1 to 0.7. This negligible fluctuation confirms that DDE effectively filters out corrupted samples, ensuring reliable adaptation even in heavily noisy environments.

\noindent\textbf{Analysis of Different Backbones.} As shown in~\cref{fig:backbones}, our method scales effectively across various architectures, with performance improving consistently as model capacity increases. ViT-L achieves the highest $\text{Acc}_H$ of 81.20\%, outperforming the RN50 baseline by 10.12\%. 

\noindent\textbf{Performance Stability.} \cref{fig:stability} shows that DDE maintains consistent performance across all adaptation stages. The stable $\text{Acc}_H$ across segments confirms that our method keeps performance stable over time.

\noindent\textbf{Comparison with Fixed Threshold.}
As shown in \cref{fig:fixed_threshold}, the model performance is highly sensitive to the choice of fixed thresholds. In contrast, our adaptive mechanism achieves competitive results without manual searching, nearly matching the optimal fixed parameter and demonstrating its effectiveness.

\noindent\textbf{Cache Size $Q$.} We analyze $Q$ in \cref{fig:Q}. While performance peaks at $Q=5000$ due to the increased diversity of candidate samples, maintaining such a large cache incurs substantial memory overhead. To strike an optimal balance between adaptation accuracy and memory efficiency, we set $Q=1000$ by default. 

\noindent\textbf{The Scaling Factor $\rho$.} 
We analyze $\rho$ in \cref{fig:rho}. $\text{Acc}_H$ remains relatively stable across different $\rho$ values, reaching its maximum of 78.00\% at $\rho = 0.005$. The performance remains robust even at a very small $\rho$ of 0.0005 (77.68\%).

\noindent\textbf{The Weights of Exclusion GDA Model $\beta$.} 
We analyze $\beta$ in \cref{fig:beta}. $\text{Acc}_H$ gradually improves from 77.42\% to its peak of 78.00\% as $\beta$ increases to 0.5. The performance stays consistently high across the tested range, showing the method's insensitivity to $\beta$.

\noindent\textbf{The Number of Discriminative Negative Labels $\hat{M}$.} 
We analyze $\hat{M}$ in \cref{fig:M}. The best $\text{Acc}_H$ of 78.00\% is achieved at $\hat{M} = 500$. Either too small or too large values of $\hat{M}$ significantly degrade the performance, with $\text{Acc}_H$ dropping to 71.28\% at $\hat{M} = 1000$.

\noindent\textbf{Robustness to Diverse OOD Settings.}
We further evaluate DDE under four challenging OOD settings, including heterogeneous, noisy, medical, and industrial scenarios. As shown in Table~\ref{tab:diverse_ood}, DDE consistently achieves the best performance across all settings, demonstrating strong robustness under diverse real-world OOD streams.

\noindent\textbf{Complexity Analyses.}
\Cref{tab:complexity} summarizes the efficiency of DDE. Compared with optimization-based methods such as TPT~\cite{shu2022test}, DDE achieves the lowest test-time latency (1.84 min) while introducing no additional learnable parameters and maintaining low memory consumption, demonstrating excellent efficiency and scalability.


\begin{table*}[h]
\centering

\begin{minipage}[t]{0.46\textwidth}
\centering
\small
\caption{Performance ($ACC_H$) under diverse OOD scenarios.}
\label{tab:diverse_ood}
\begin{tabular}{l|c|c|c|c}
\hline
\textbf{Methods}& Heter & Noise & Medical & Industrial \\
\hline
CLIP & 63.23 & 65.92 & 33.45 & 60.63 \\
AdaND & 66.25 & 67.03 & 36.82 & 64.25 \\
\textbf{DDE} & \textbf{74.89} & \textbf{75.90} & \textbf{45.96} & \textbf{71.15} \\
\hline
\end{tabular}
\end{minipage}
\hfill
\begin{minipage}[t]{0.48\textwidth}
\centering
\scriptsize
\setlength{\tabcolsep}{2pt}
\caption{Complexity analyses. Results are obtained using a GeForce RTX 3090.}
\label{tab:complexity}
\begin{tabular}{lccccc} 
\toprule
\textbf{Methods} &
\textbf{Testing(min)} &
\textbf{Memory(GiB)} &
\textbf{FPS$\uparrow$} &
\textbf{Param(k).} &
\textbf{$\text{Acc}_H \uparrow$} \\
\midrule
TENT~\cite{wang2020tent}
& 3.72 & 14.99 & 270 & 40 & 63.95 \\
TPT~\cite{shu2022test}
& 321.90 & 21.23 & 3.11 & 8.2 & 64.77 \\
AdaNeg~\cite{zhang2024adaneg}
& 2.10 & 5.90 & 476 & -- & 73.76 \\
AdaND~\cite{cao2025noisy}
& 6.59 & 4.57 & 285 & 1.0 & 72.77 \\
\rowcolor{gray!15}
\textbf{DDE}
& 1.84 & 5.41 & 545 & -- & 78.00 \\
\bottomrule
\end{tabular}
\end{minipage}
\end{table*}

\vspace{-1.5cm}
\section{Conclusion}
We presented Dual Distribution Estimation (DDE), a novel, training-free, zero-shot framework tailored for NTTA. We first identified several limitations inherent in existing NTTA methods. To address these pitfalls, we incorporated two key components: (1) Positive Feature Distribution Estimation, which utilized dual Gaussian distributions (inclusion and exclusion) to refine ID accuracy; and (2) Negative Label Distribution Estimation, which selected discriminative negative labels to filter out noise from generic ones. Extensive experiments on large-scale benchmarks demonstrated that DDE achieved state-of-the-art performance while maintaining both robustness and efficiency.
\noindent One minor limitation of DDE is maintaining class-wise Gaussian parameters introduces a marginal memory overhead on extremely large-scale datasets. It is generally negligible in typical scenarios and does not detract from the overall efficiency and effectiveness of DDE.


{\small
\bibliography{ref}

\begin{thebibliography}{69}
\providecommand{\natexlab}[1]{#1}
\providecommand{\url}[1]{\texttt{#1}}
\expandafter\ifx\csname urlstyle\endcsname\relax
  \providecommand{\doi}[1]{doi: #1}\else
  \providecommand{\doi}{doi: \begingroup \urlstyle{rm}\Url}\fi

\bibitem[Cao et~al.(2025)Cao, Zhong, Zhou, Liu, Liu, Zhang, and Han]{cao2025noisy}
Chentao Cao, Zhun Zhong, Zhanke Zhou, Tongliang Liu, Yang Liu, Kun Zhang, and Bo~Han.
\newblock Noisy test-time adaptation in vision-language models.
\newblock \emph{arXiv preprint arXiv:2502.14604}, 2025.

\bibitem[Zhang et~al.(2024{\natexlab{a}})Zhang, Zhu, Tang, Ma, Zhou, and Zhang]{zhang2024dual_cvpr}
Yabin Zhang, Wenjie Zhu, Hui Tang, Zhiyuan Ma, Kaiyang Zhou, and Lei Zhang.
\newblock Dual memory networks: A versatile adaptation approach for vision-language models.
\newblock In \emph{Proceedings of the IEEE/CVF conference on computer vision and pattern recognition}, pages 28718--28728, 2024{\natexlab{a}}.

\bibitem[Karmanov et~al.(2024)Karmanov, Guan, Lu, El~Saddik, and Xing]{karmanov2024efficient}
Adilbek Karmanov, Dayan Guan, Shijian Lu, Abdulmotaleb El~Saddik, and Eric Xing.
\newblock Efficient test-time adaptation of vision-language models.
\newblock In \emph{Proceedings of the IEEE/CVF Conference on Computer Vision and Pattern Recognition}, pages 14162--14171, 2024.

\bibitem[Chen et~al.(2025)Chen, Zhai, Zhang, Shi, and Li]{chen2025multi}
Xinyu Chen, Haotian Zhai, Can Zhang, Xiupeng Shi, and Ruirui Li.
\newblock Multi-cache enhanced prototype learning for test-time generalization of vision-language models.
\newblock In \emph{Proceedings of the IEEE/CVF International Conference on Computer Vision}, pages 2281--2291, 2025.

\bibitem[Zhang et~al.(2024{\natexlab{b}})Zhang, Stepputtis, Sycara, and Xie]{zhang2024dual_neurips}
Ce~Zhang, Simon Stepputtis, Katia Sycara, and Yaqi Xie.
\newblock Dual prototype evolving for test-time generalization of vision-language models.
\newblock \emph{Advances in Neural Information Processing Systems}, 37:\penalty0 32111--32136, 2024{\natexlab{b}}.

\bibitem[Zhang and Zhang(2024)]{zhang2024adaneg}
Yabin Zhang and Lei Zhang.
\newblock Adaneg: Adaptive negative proxy guided ood detection with vision-language models.
\newblock \emph{Advances in Neural Information Processing Systems}, 37:\penalty0 38744--38768, 2024.

\bibitem[Yang et~al.(2025)Yang, Zhu, Sun, Liu, Gu, and Ye]{yang2025oodd}
Yifeng Yang, Lin Zhu, Zewen Sun, Hengyu Liu, Qinying Gu, and Nanyang Ye.
\newblock Oodd: Test-time out-of-distribution detection with dynamic dictionary.
\newblock In \emph{Proceedings of the Computer Vision and Pattern Recognition Conference}, pages 30630--30639, 2025.

\bibitem[Li et~al.(2023)Li, Xu, Su, and Jia]{li2023robustness}
Yushu Li, Xun Xu, Yongyi Su, and Kui Jia.
\newblock On the robustness of open-world test-time training: Self-training with dynamic prototype expansion.
\newblock In \emph{Proceedings of the IEEE/CVF International Conference on Computer Vision}, pages 11836--11846, 2023.

\bibitem[Wang et~al.(2020)Wang, Shelhamer, Liu, Olshausen, and Darrell]{wang2020tent}
Dequan Wang, Evan Shelhamer, Shaoteng Liu, Bruno Olshausen, and Trevor Darrell.
\newblock Tent: Fully test-time adaptation by entropy minimization.
\newblock \emph{arXiv preprint arXiv:2006.10726}, 2020.

\bibitem[Niu et~al.(2023)Niu, Wu, Zhang, Wen, Chen, Zhao, and Tan]{niu2023towards}
Shuaicheng Niu, Jiaxiang Wu, Yifan Zhang, Zhiquan Wen, Yaofo Chen, Peilin Zhao, and Mingkui Tan.
\newblock Towards stable test-time adaptation in dynamic wild world.
\newblock \emph{arXiv preprint arXiv:2302.12400}, 2023.

\bibitem[Shi et~al.(2022)Shi, Zhang, Tang, Zhu, Li, Guo, and Zhuang]{shi2022efficacy}
Haizhou Shi, Youcai Zhang, Siliang Tang, Wenjie Zhu, Yaqian Li, Yandong Guo, and Yueting Zhuang.
\newblock On the efficacy of small self-supervised contrastive models without distillation signals.
\newblock In \emph{Proceedings of the AAAI conference on artificial intelligence}, volume~36, pages 2225--2234, 2022.

\bibitem[Meng et~al.(2026)Meng, Gu, Liang, Lalor, Chambers, and Chen]{meng2026topocl}
Guangyu Meng, Pengfei Gu, Peixian Liang, John~P Lalor, Erin~Wolf Chambers, and Danny~Z Chen.
\newblock Topocl: Topological contrastive learning for medical imaging.
\newblock In \emph{Proceedings of the IEEE/CVF Conference on Computer Vision and Pattern Recognition}, pages 42681--42690, 2026.

\bibitem[Ma et~al.(2025)Ma, Chen, Zhang, Wu, and Ding]{ma2025instruct}
Weijian Ma, Ruoxin Chen, Keyue Zhang, Shuang Wu, and Shouhong Ding.
\newblock Instruct where the model fails: Generative data augmentation via guided self-contrastive fine-tuning.
\newblock In \emph{Proceedings of the AAAI Conference on Artificial Intelligence}, volume~39, pages 5991--5999, 2025.

\bibitem[Mirza et~al.(2022)Mirza, Micorek, Possegger, and Bischof]{mirza2022norm}
M~Jehanzeb Mirza, Jakub Micorek, Horst Possegger, and Horst Bischof.
\newblock The norm must go on: Dynamic unsupervised domain adaptation by normalization.
\newblock In \emph{Proceedings of the IEEE/CVF conference on computer vision and pattern recognition}, pages 14765--14775, 2022.

\bibitem[Zhao et~al.(2023)Zhao, Chen, and Xia]{zhao2023delta}
Bowen Zhao, Chen Chen, and Shu-Tao Xia.
\newblock Delta: degradation-free fully test-time adaptation.
\newblock \emph{arXiv preprint arXiv:2301.13018}, 2023.

\bibitem[Sui et~al.(2025)Sui, Wang, and Yeung-Levy]{sui2025just}
Elaine Sui, Xiaohan Wang, and Serena Yeung-Levy.
\newblock Just shift it: Test-time prototype shifting for zero-shot generalization with vision-language models.
\newblock In \emph{2025 IEEE/CVF Winter Conference on Applications of Computer Vision (WACV)}, pages 825--835. IEEE, 2025.

\bibitem[Zhu et~al.(2024{\natexlab{a}})Zhu, Zhu, Tan, Wang, Hao, and Zhang]{zhu2024enhancing}
Xingyu Zhu, Beier Zhu, Yi~Tan, Shuo Wang, Yanbin Hao, and Hanwang Zhang.
\newblock Enhancing zero-shot vision models by label-free prompt distribution learning and bias correcting.
\newblock \emph{Advances in Neural Information Processing Systems}, 37:\penalty0 2001--2025, 2024{\natexlab{a}}.

\bibitem[Han et~al.(2024)Han, Yang, Wang, Li, Xu, Shou, and Zhang]{han2024dota}
Zongbo Han, Jialong Yang, Guangyu Wang, Junfan Li, Qianli Xu, Mike~Zheng Shou, and Changqing Zhang.
\newblock Dota: Distributional test-time adaptation of vision-language models.
\newblock \emph{arXiv preprint arXiv:2409.19375}, 2024.

\bibitem[Lafon et~al.(2025)Lafon, Hakim, Rambour, Desrosier, and Thome]{lafon2025cliptta}
Marc Lafon, Gustavo Adolfo~Vargas Hakim, Cl{\'e}ment Rambour, Christian Desrosier, and Nicolas Thome.
\newblock Cliptta: Robust contrastive vision-language test-time adaptation.
\newblock \emph{arXiv preprint arXiv:2507.14312}, 2025.

\bibitem[Zhou et~al.(2025)Zhou, Ye, Li, Li, Zhu, Deng, Liu, and Lei]{zhou2025bayesian}
Lihua Zhou, Mao Ye, Shuaifeng Li, Nianxin Li, Xiatian Zhu, Lei Deng, Hongbin Liu, and Zhen Lei.
\newblock Bayesian test-time adaptation for vision-language models.
\newblock In \emph{Proceedings of the Computer Vision and Pattern Recognition Conference}, pages 29999--30009, 2025.

\bibitem[Luo et~al.(2026)Luo, Ou, Deng, Deng, Yan, Wen, and Tan]{luo2026protodcs}
Wei Luo, Yangfan Ou, Jin Deng, Zeshuai Deng, Xiquan Yan, Zhiquan Wen, and Mingkui Tan.
\newblock Protodcs: Towards robust and efficient open-set test-time adaptation for vision-language models.
\newblock \emph{arXiv preprint arXiv:2602.23653}, 2026.

\bibitem[Zhang et~al.(2025)Zhang, Wang, Zhou, Yuan, Zhang, Wang, and Jin]{zhang2025model}
YiFan Zhang, Xue Wang, Tian Zhou, Kun Yuan, Zhang Zhang, Liang Wang, and Rong Jin.
\newblock Model-free test time adaptation for out-of-distribution detection.
\newblock \emph{IEEE Transactions on Pattern Analysis and Machine Intelligence}, 2025.

\bibitem[Yang et~al.(2023)Yang, Akhtar, Wen, Shah, and Mian]{yang2023re}
Peiyu Yang, Naveed Akhtar, Zeyi Wen, Mubarak Shah, and Ajmal~Saeed Mian.
\newblock Re-calibrating feature attributions for model interpretation.
\newblock In \emph{International Conference on Learning Representations}, 2023.

\bibitem[Yang et~al.(2024{\natexlab{a}})Yang, Akhtar, Jiang, and Mian]{yang2024backdoor}
Peiyu Yang, Naveed Akhtar, Jiantong Jiang, and Ajmal Mian.
\newblock Backdoor-based explainable ai benchmark for high fidelity evaluation of attribution methods.
\newblock \emph{arXiv preprint arXiv:2405.02344}, 2024{\natexlab{a}}.

\bibitem[Lu et~al.(2025)Lu, Wang, Sheng, He, Zheng, and Liang]{lu2025out}
Shuo Lu, Yingsheng Wang, Lijun Sheng, Lingxiao He, Aihua Zheng, and Jian Liang.
\newblock Out-of-distribution detection: A task-oriented survey of recent advances.
\newblock \emph{ACM Computing Surveys}, 58\penalty0 (2):\penalty0 1--39, 2025.

\bibitem[Feng et~al.(2026{\natexlab{a}})Feng, Jiang, Zhou, and Ge]{feng2026beyond}
Wei Feng, Yiwen Jiang, Sijin Zhou, and Zongyuan Ge.
\newblock Beyond the static world: Continual category discovery under visual drift.
\newblock In \emph{Proceedings of the IEEE/CVF Conference on Computer Vision and Pattern Recognition}, pages 25032--25042, 2026{\natexlab{a}}.

\bibitem[Feng et~al.(2026{\natexlab{b}})Feng, Jiang, Zhou, Qi, Xu, Wang, Tang, and Ge]{feng2026seeing}
Wei Feng, Yiwen Jiang, Sijin Zhou, Zhuang Qi, Zhongxing Xu, Zhonghua Wang, Feilong Tang, and Zongyuan Ge.
\newblock Seeing through the shift: Causality-inspired robust generalized category discovery.
\newblock In \emph{Proceedings of the IEEE/CVF Conference on Computer Vision and Pattern Recognition}, pages 17766--17775, 2026{\natexlab{b}}.

\bibitem[Feng and Ge(2026)]{feng2026generalized}
Wei Feng and Zongyuan Ge.
\newblock Generalized category discovery under domain shift: A frequency domain perspective.
\newblock \emph{Advances in Neural Information Processing Systems}, 38:\penalty0 111721--111749, 2026.

\bibitem[Ning et~al.(2024)Ning, Huang, Ding, Wang, and Zhu]{ning2024physics}
Jia Ning, Weiguo Huang, Chuancang Ding, Jun Wang, and Zhongkui Zhu.
\newblock Physics-informed unsupervised domain adaptation framework for cross-machine bearing fault diagnosis.
\newblock \emph{Advanced Engineering Informatics}, 62:\penalty0 102774, 2024.

\bibitem[Ning et~al.(2025)Ning, Huang, Guo, Ding, Huangfu, Shen, and Zhu]{ning2025physics}
Jia Ning, Weiguo Huang, Panpan Guo, Chuancang Ding, Yifan Huangfu, Changqing Shen, and Zhongkui Zhu.
\newblock A physics-guided memory enhancement and causality-inspired generalization framework for continual fault diagnosis.
\newblock \emph{Knowledge-Based Systems}, 325:\penalty0 114044, 2025.
\newblock Corresponding author: Weiguo Huang.

\bibitem[He et~al.(2026)He, Zhang, Chen, and Cao]{he2026cinematte}
Yuanjian He, Chen Zhang, Fasheng Chen, and Jiangbo Cao.
\newblock Cinematte: Background matting for virtual production and beyond.
\newblock In \emph{Proceedings of the IEEE/CVF Conference on Computer Vision and Pattern Recognition}, pages 8725--8735, 2026.

\bibitem[Zhu et~al.(2024{\natexlab{b}})Zhu, Lin, Tan, Zhu, Li, Wang, and Li]{zhu2024advancing}
Chunzheng Zhu, Jianxin Lin, Guanghua Tan, Ningbo Zhu, Kenli Li, Chunlian Wang, and Shengli Li.
\newblock Advancing ultrasound medical continuous learning with task-specific generalization and adaptability.
\newblock In \emph{2024 IEEE International Conference on Bioinformatics and Biomedicine (BIBM)}, pages 3019--3025. IEEE, 2024{\natexlab{b}}.

\bibitem[Zhu et~al.(2026{\natexlab{a}})Zhu, Lin, Chen, Wang, and Lin]{zhu2026medeyes}
Chunzheng Zhu, Yangfang Lin, Shen Chen, Yijun Wang, and Jianxin Lin.
\newblock Medeyes: Learning dynamic visual focus for medical progressive diagnosis.
\newblock In \emph{Proceedings of the AAAI Conference on Artificial Intelligence}, volume~40, pages 13916--13924, 2026{\natexlab{a}}.

\bibitem[Tao et~al.(2026)Tao, Wang, Song, Luo, Liu, and Liu]{tao2026grasp}
Yicheng Tao, Yiqun Wang, Xiangchen Song, Xin Luo, Kai Liu, and Jie Liu.
\newblock Grasp: Plan-guided graph retrieval with adaptive fusion and reranking on semi-structured knowledge bases.
\newblock \emph{arXiv preprint arXiv:2605.30237}, 2026.

\bibitem[Fu et~al.(2026)Fu, Tang, Wang, Tan, Zhang, Kang, Qi, Zhang, and Fong]{fu2026modalimmune}
Rong Fu, WeiZhi Tang, Ziming Wang, Jia~Yee Tan, Zijian Zhang, Zhaolu Kang, Muge Qi, Shuning Zhang, and Simon Fong.
\newblock Modalimmune: Immunity driven unlearning via self destructive training.
\newblock \emph{arXiv preprint arXiv:2602.16197}, 2026.

\bibitem[Wang et~al.(2026)Wang, Li, Li, Chen, Huang, Chen, Li, Liu, and Chen]{wang2026sppo}
Tianyi Wang, Yixia Li, Long Li, Yibiao Chen, Shaohan Huang, Yun Chen, Peng Li, Yang Liu, and Guanhua Chen.
\newblock Sppo: Sequence-level ppo for long-horizon reasoning tasks, 2026.
\newblock URL \url{https://arxiv.org/abs/2604.08865}.

\bibitem[Yang et~al.(2024{\natexlab{b}})Yang, Akhtar, Shah, and Mian]{yang2024regulating}
Peiyu Yang, Naveed Akhtar, Mubarak Shah, and Ajmal Mian.
\newblock Regulating model reliance on non-robust features by smoothing input marginal density.
\newblock In \emph{European Conference on Computer Vision}, pages 329--347. Springer, 2024{\natexlab{b}}.

\bibitem[Zhang et~al.(2024{\natexlab{c}})Zhang, Zhu, He, and Zhang]{zhang2024lapt}
Yabin Zhang, Wenjie Zhu, Chenhang He, and Lei Zhang.
\newblock Lapt: Label-driven automated prompt tuning for ood detection with vision-language models.
\newblock In \emph{European conference on computer vision}, pages 271--288. Springer, 2024{\natexlab{c}}.

\bibitem[Zhu et~al.(2025)Zhu, Zhang, Jin, Zeng, and Zhang]{zhu2025knowledge}
Wenjie Zhu, Yabin Zhang, Xin Jin, Wenjun Zeng, and Lei Zhang.
\newblock Knowledge regularized negative feature tuning of vision-language models for out-of-distribution detection.
\newblock In \emph{Proceedings of the 33rd ACM International Conference on Multimedia}, pages 3565--3574, 2025.

\bibitem[Zhu et~al.(2026{\natexlab{b}})Zhu, Zhang, Jin, Zeng, and Zhang]{zhu2026ants}
Wenjie Zhu, Yabin Zhang, Xin Jin, Wenjun Zeng, and Lei Zhang.
\newblock Ants: Adaptive negative textual space shaping for ood detection via test-time mllm understanding and reasoning.
\newblock In \emph{Proceedings of the IEEE/CVF Conference on Computer Vision and Pattern Recognition}, pages 20--30, 2026{\natexlab{b}}.

\bibitem[Zhang et~al.(2026{\natexlab{a}})Zhang, Varma, Gao, Delbrouck, Liu, Wang, and Langlotz]{zhang2026activation}
Yabin Zhang, Maya Varma, Yunhe Gao, Jean-Benoit Delbrouck, Jiaming Liu, Chong Wang, and Curtis Langlotz.
\newblock Activation matters: Test-time activated negative labels for ood detection with vision-language models.
\newblock \emph{arXiv preprint arXiv:2603.25250}, 2026{\natexlab{a}}.

\bibitem[Tang et~al.(2026)Tang, Liu, Yan, Shen, He, and Qin]{tang2026cross}
Hao Tang, Yu~Liu, Shuanglin Yan, Fei Shen, Shengfeng He, and Jing Qin.
\newblock Cross-modal proxy evolving for ood detection with vision-language models.
\newblock \emph{arXiv preprint arXiv:2601.08476}, 2026.

\bibitem[Guo et~al.(2025)Guo, Luo, Zheng, Wang, Chang, Wang, and Liu]{guo2025quantized}
Jiajun Guo, Xin Luo, Jiayin Zheng, Yiqun Wang, Kai-Wei Chang, Wei Wang, and Jie Liu.
\newblock Quantized-tinyllava: a new multimodal foundation model enables efficient split learning.
\newblock \emph{arXiv preprint arXiv:2511.23402}, 2025.

\bibitem[Zeng et~al.(2025)Zeng, Chang, Xie, Liu, Bai, Pan, Xu, and Wei]{zeng2025FSDrive}
Shuang Zeng, Xinyuan Chang, Mengwei Xie, Xinran Liu, Yifan Bai, Zheng Pan, Mu~Xu, and Xing Wei.
\newblock Futuresightdrive: Thinking visually with spatio-temporal cot for autonomous driving.
\newblock \emph{arXiv preprint arXiv:2505.17685}, 2025.

\bibitem[Xiao et~al.(2026)Xiao, Xu, Ma, Jiang, Gao, and Wu]{xiao2026reversible}
Canran Xiao, Tianxiang Xu, Siyuan Ma, Yiyang Jiang, Haoyu Gao, and Yuhan Wu.
\newblock Reversible primitive--composition alignment for continual vision--language learning.
\newblock In \emph{The Fourteenth International Conference on Learning Representations}, 2026.

\bibitem[Zhang et~al.(2026{\natexlab{b}})Zhang, Zhao, Xiao, Duan, Mo, Gao, and Wang]{zhang2026pi}
Jiayu Zhang, Chuangxin Zhao, Canran Xiao, Ruibo Duan, Wenyi Mo, Haoyu Gao, and Wenshuo Wang.
\newblock Pi-cca: Prompt-invariant cca certificates for replay-free continual multimodal learning.
\newblock In \emph{The Fourteenth International Conference on Learning Representations}, 2026{\natexlab{b}}.

\bibitem[Ma et~al.(2026)Ma, Sun, Yu, Wang, Chua, and Bian]{ma2026thinking}
Weijian Ma, Shizhao Sun, Tianyu Yu, Ruiyu Wang, Tat-Seng Chua, and Jiang Bian.
\newblock Thinking with blueprints: Assisting vision-language models in spatial reasoning via structured object representation.
\newblock \emph{arXiv preprint arXiv:2601.01984}, 2026.

\bibitem[Li et~al.(2025{\natexlab{a}})Li, Dong, Yang, Wen, Koniusz, Huang, Tian, and Ong]{li2025mmt}
Yuqi Li, Junhao Dong, Chuanguang Yang, Shiping Wen, Piotr Koniusz, Tingwen Huang, Yingli Tian, and Yew-Soon Ong.
\newblock Mmt-ard: Multimodal multi-teacher adversarial distillation for robust vision-language models.
\newblock \emph{arXiv preprint arXiv:2511.17448}, 2025{\natexlab{a}}.

\bibitem[Li et~al.(2025{\natexlab{b}})Li, Yang, Dong, Yao, Xu, Dong, Zeng, An, and Tian]{li2025ammkd}
Yuqi Li, Chuanguang Yang, Junhao Dong, Zhengtao Yao, Haoyan Xu, Zeyu Dong, Hansheng Zeng, Zhulin An, and Yingli Tian.
\newblock Ammkd: Adaptive multimodal multi-teacher distillation for lightweight vision-language models.
\newblock \emph{arXiv preprint arXiv:2509.00039}, 2025{\natexlab{b}}.

\bibitem[Hastie and Tibshirani(1996)]{hastie1996discriminant}
Trevor Hastie and Robert Tibshirani.
\newblock Discriminant analysis by gaussian mixtures.
\newblock \emph{Journal of the Royal Statistical Society Series B: Statistical Methodology}, 58\penalty0 (1):\penalty0 155--176, 1996.

\bibitem[Ming et~al.(2022)Ming, Cai, Gu, Sun, Li, and Li]{ming2022delving}
Yifei Ming, Ziyang Cai, Jiuxiang Gu, Yiyou Sun, Wei Li, and Yixuan Li.
\newblock Delving into out-of-distribution detection with vision-language representations.
\newblock \emph{Advances in neural information processing systems}, 35:\penalty0 35087--35102, 2022.

\bibitem[Jiang et~al.(2024)Jiang, Liu, Fang, Chen, Liu, Zheng, and Han]{jiang2024negative}
Xue Jiang, Feng Liu, Zhen Fang, Hong Chen, Tongliang Liu, Feng Zheng, and Bo~Han.
\newblock Negative label guided ood detection with pretrained vision-language models.
\newblock \emph{arXiv preprint arXiv:2403.20078}, 2024.

\bibitem[Bishop and Nasrabadi(2006)]{bishop2006pattern}
Christopher~M Bishop and Nasser~M Nasrabadi.
\newblock \emph{Pattern recognition and machine learning}, volume~4.
\newblock Springer, 2006.

\bibitem[Friedman(1989)]{friedman1989regularized}
Jerome~H Friedman.
\newblock Regularized discriminant analysis.
\newblock \emph{Journal of the American statistical association}, 84\penalty0 (405):\penalty0 165--175, 1989.

\bibitem[Radford et~al.(2021)Radford, Kim, Hallacy, Ramesh, Goh, Agarwal, Sastry, Askell, Mishkin, Clark, et~al.]{radford2021learning}
Alec Radford, Jong~Wook Kim, Chris Hallacy, Aditya Ramesh, Gabriel Goh, Sandhini Agarwal, Girish Sastry, Amanda Askell, Pamela Mishkin, Jack Clark, et~al.
\newblock Learning transferable visual models from natural language supervision.
\newblock In \emph{International conference on machine learning}, pages 8748--8763. PmLR, 2021.

\bibitem[Shu et~al.(2022)Shu, Nie, Huang, Yu, Goldstein, Anandkumar, and Xiao]{shu2022test}
Manli Shu, Weili Nie, De-An Huang, Zhiding Yu, Tom Goldstein, Anima Anandkumar, and Chaowei Xiao.
\newblock Test-time prompt tuning for zero-shot generalization in vision-language models.
\newblock \emph{Advances in Neural Information Processing Systems}, 35:\penalty0 14274--14289, 2022.

\bibitem[Deng et~al.(2009)Deng, Dong, Socher, Li, Li, and Fei-Fei]{deng2009imagenet}
Jia Deng, Wei Dong, Richard Socher, Li-Jia Li, Kai Li, and Li~Fei-Fei.
\newblock Imagenet: A large-scale hierarchical image database.
\newblock In \emph{2009 IEEE conference on computer vision and pattern recognition}, pages 248--255. Ieee, 2009.

\bibitem[Wang et~al.(2019)Wang, Ge, Lipton, and Xing]{wang2019learning}
Haohan Wang, Songwei Ge, Zachary Lipton, and Eric~P Xing.
\newblock Learning robust global representations by penalizing local predictive power.
\newblock \emph{Advances in neural information processing systems}, 32, 2019.

\bibitem[Hendrycks et~al.(2021{\natexlab{a}})Hendrycks, Zhao, Basart, Steinhardt, and Song]{hendrycks2021natural}
Dan Hendrycks, Kevin Zhao, Steven Basart, Jacob Steinhardt, and Dawn Song.
\newblock Natural adversarial examples.
\newblock In \emph{Proceedings of the IEEE/CVF conference on computer vision and pattern recognition}, pages 15262--15271, 2021{\natexlab{a}}.

\bibitem[Recht et~al.(2019)Recht, Roelofs, Schmidt, and Shankar]{recht2019imagenet}
Benjamin Recht, Rebecca Roelofs, Ludwig Schmidt, and Vaishaal Shankar.
\newblock Do imagenet classifiers generalize to imagenet?
\newblock In \emph{International conference on machine learning}, pages 5389--5400. PMLR, 2019.

\bibitem[Hendrycks et~al.(2021{\natexlab{b}})Hendrycks, Basart, Mu, Kadavath, Wang, Dorundo, Desai, Zhu, Parajuli, Guo, et~al.]{hendrycks2021many}
Dan Hendrycks, Steven Basart, Norman Mu, Saurav Kadavath, Frank Wang, Evan Dorundo, Rahul Desai, Tyler Zhu, Samyak Parajuli, Mike Guo, et~al.
\newblock The many faces of robustness: A critical analysis of out-of-distribution generalization.
\newblock In \emph{Proceedings of the IEEE/CVF international conference on computer vision}, pages 8340--8349, 2021{\natexlab{b}}.

\bibitem[Wah et~al.()Wah, Branson, Welinder, Perona, Belongie, et~al.]{wah2011caltech}
Catherine Wah, Steve Branson, Peter Welinder, Pietro Perona, Serge Belongie, et~al.
\newblock The caltech-ucsd birds-200-2011 dataset.
\newblock Technical report.

\bibitem[Krause et~al.(2013)Krause, Stark, Deng, and Fei-Fei]{krause20133d}
Jonathan Krause, Michael Stark, Jia Deng, and Li~Fei-Fei.
\newblock 3d object representations for fine-grained categorization.
\newblock In \emph{Proceedings of the IEEE international conference on computer vision workshops}, pages 554--561, 2013.

\bibitem[Bossard et~al.(2014)Bossard, Guillaumin, and Van~Gool]{bossard2014food}
Lukas Bossard, Matthieu Guillaumin, and Luc Van~Gool.
\newblock Food-101--mining discriminative components with random forests.
\newblock In \emph{European conference on computer vision}, pages 446--461. Springer, 2014.

\bibitem[Parkhi et~al.(2012)Parkhi, Vedaldi, Zisserman, and Jawahar]{parkhi2012cats}
Omkar~M Parkhi, Andrea Vedaldi, Andrew Zisserman, and CV~Jawahar.
\newblock Cats and dogs.
\newblock In \emph{2012 IEEE conference on computer vision and pattern recognition}, pages 3498--3505. IEEE, 2012.

\bibitem[Van~Horn et~al.(2018)Van~Horn, Mac~Aodha, Song, Cui, Sun, Shepard, Adam, Perona, and Belongie]{van2018inaturalist}
Grant Van~Horn, Oisin Mac~Aodha, Yang Song, Yin Cui, Chen Sun, Alex Shepard, Hartwig Adam, Pietro Perona, and Serge Belongie.
\newblock The inaturalist species classification and detection dataset.
\newblock In \emph{Proceedings of the IEEE conference on computer vision and pattern recognition}, pages 8769--8778, 2018.

\bibitem[Xiao et~al.(2010)Xiao, Hays, Ehinger, Oliva, and Torralba]{xiao2010sun}
Jianxiong Xiao, James Hays, Krista~A Ehinger, Aude Oliva, and Antonio Torralba.
\newblock Sun database: Large-scale scene recognition from abbey to zoo.
\newblock In \emph{2010 IEEE computer society conference on computer vision and pattern recognition}, pages 3485--3492. IEEE, 2010.

\bibitem[Cimpoi et~al.(2014)Cimpoi, Maji, Kokkinos, Mohamed, and Vedaldi]{cimpoi2014describing}
Mircea Cimpoi, Subhransu Maji, Iasonas Kokkinos, Sammy Mohamed, and Andrea Vedaldi.
\newblock Describing textures in the wild.
\newblock In \emph{Proceedings of the IEEE conference on computer vision and pattern recognition}, pages 3606--3613, 2014.

\bibitem[Zhou et~al.(2017)Zhou, Lapedriza, Khosla, Oliva, and Torralba]{zhou2017places}
Bolei Zhou, Agata Lapedriza, Aditya Khosla, Aude Oliva, and Antonio Torralba.
\newblock Places: A 10 million image database for scene recognition.
\newblock \emph{IEEE transactions on pattern analysis and machine intelligence}, 40\penalty0 (6):\penalty0 1452--1464, 2017.

\end{thebibliography}
}

\appendix

\clearpage
\setcounter{table}{0}
\setcounter{equation}{0}
\setcounter{figure}{0}
\renewcommand{\thetable}{\thesection.\arabic{table}}
\renewcommand{\theequation}{\thesection.\arabic{equation}}
\renewcommand{\thefigure}{\thesection.\arabic{figure}}
\begin{center}
    {\LARGE\bfseries Supplementary Material}
\end{center}



\renewcommand{\thetable}{\Alph{table}}
\renewcommand{\thefigure}{\Alph{figure}}
\renewcommand{\thefigure}{S\arabic{figure}}
\renewcommand{\thetable}{S\arabic{table}}
\renewcommand{\theequation}{S\arabic{equation}}
\setcounter{equation}{0}

\begin{table}[h]
\centering
\caption{Hyper-parameter settings for the proposed ZS-NTTA framework.}
\label{tab:hyperparameters}
\begin{tabular}{lcc}
\toprule
\textbf{Description} & \textbf{Parameter} & \textbf{Value} \\ 
\midrule
\multicolumn{3}{l}{\textit{Positive and Negative Images Selection}} \\
Negative labels & $M$ & 10000 \\
Group number & $g$ & 5 \\
Positive threshold & $\lambda_{pos}$ & 0.75 \\
Negative threshold & $\lambda_{neg}$ & 0.25 \\
Queue length & $Q$ & 1000 \\
\midrule
\multicolumn{3}{l}{\textit{Positive Feature Distribution Estimation}} \\
Exclusion GDA weight & $\beta$ & 0.5 \\
Scaling factor & $\rho$ & 0.005 \\
Batch size & $B$ & 128 \\
Maximum GDA weight & $\alpha_{max}$ & 1.0 \\
\midrule
\multicolumn{3}{l}{\textit{Negative Label Distribution Estimation}} \\
Discriminative negative labels & $\hat{M}$ & 500 \\
\bottomrule
\end{tabular}
\end{table}

\noindent\textbf{Experimental Details.}
\cref{tab:hyperparameters} outlines the specific hyper-parameter configurations utilized across our experimental framework. For the selection of positive and negative anchors, we maintain a comprehensive pool of $M=10000$ negative labels and a queue of length $Q=1000$ to ensure statistical diversity. The thresholds $\lambda_{pos}$ and $\lambda_{neg}$ are fixed at 0.75 and 0.25, respectively, to maintain high-precision sample filtering. Furthermore, to facilitate the instantialization of the exclusion distribution, the exclusion GDA weight $\beta$ is set to 0.5, striking a balance between distribution calibration and feature robustness during the test-time adaptation process.

\noindent\textbf{Evaluation Metric.}
To assess the performance under the ZS-NTTA framework, we employ three primary metrics: $Acc_{S}$, $Acc_{N}$, and $Acc_{H}$. 
Specifically, $Acc_{S}$ denotes the classification accuracy on clean (source-like) samples, while $Acc_{N}$ evaluates the model's robustness on corrupted or noisy data. 
To provide a unified reflection of the trade-off between clean-set stability and noise-set adaptation, we introduce the harmonic mean of the two, denoted as $Acc_{H}$, which is formulated as follows:

\begin{table}[h]  
\centering       
\begin{equation} 
\begin{aligned}
\text{Acc}_{\text{S}} &= \frac{\sum_{x_i, y_i \in \mathcal{D}} \mathbbm{1}(y_i = \hat{y}_i) \cdot \mathbbm{1}(y_i \in \mathcal{Y}_{\text{id}})}{\sum_{x_i, y_i \in \mathcal{D}} \mathbbm{1}(y_i \in \mathcal{Y}_{\text{id}})}, \\
\text{Acc}_{\text{N}} &= \frac{\sum_{x_i, y_i \in \mathcal{D}} \mathbbm{1}(\hat{y}_i \in \mathcal{Y}_{\text{noisy}}) \cdot \mathbbm{1}(y_i \in \mathcal{Y}_{\text{noisy}})}{\sum_{x_i, y_i \in \mathcal{D}} \mathbbm{1}(y_i \in \mathcal{Y}_{\text{noisy}})}, \\
\text{Acc}_{\text{H}} &= 2 \cdot \frac{\text{Acc}_{\text{S}} \cdot \text{Acc}_{\text{N}}}{\text{Acc}_{\text{S}} + \text{Acc}_{\text{N}}}.
\end{aligned}
\label{eq:metrics}
\end{equation}
\end{table}

\noindent\textbf{Zero-shot Noisy TTA Results for Near OOD.}
Table \ref{tab:zstta_near_ood} presents the comparative results on ImageNet-1k as the ID dataset. Our proposed DDE significantly outperforms all baseline methods across two challenging Near OOD benchmarks (SSB\_Hard and NINCO). Specifically, on SSB\_Hard, DDE achieves a remarkable balance between Seen and Noisy accuracy, reaching an H-mean of 61.53\%, which is a 4.09\% improvement over the previous best competitor, AdaNeg (57.44\%). While some methods like AdaND exhibit strong $Acc_S$ but suffer from a severe performance drop in $Acc_N$, DDE maintains robust performance in both scenarios. Overall, DDE achieves the highest average H-mean of 63.40\%, demonstrating its superior capability in handling distribution shifts and label noise simultaneously during test-time adaptation.

\noindent\begin{table}[t]
\centering
\caption{Zero-shot noisy TTA results for ImageNet-1k as the ID dataset. \textbf{Bold} indicates the best performance.}
\label{tab:zstta_near_ood}
\scriptsize
\begin{tabular}{l *{9}{c}}
\toprule
\multirow{2}{*}{\textbf{Method}} & \multicolumn{3}{c}{\textbf{SSB\_Hard}} & \multicolumn{3}{c}{\textbf{NINCO}} & \multicolumn{3}{c}{\textbf{Avg}} \\
\cmidrule(lr){2-4} \cmidrule(lr){5-7} \cmidrule(lr){8-10}
& $Acc_S$ & $Acc_N$ & $Acc_H$ & $Acc_S$ & $Acc_N$ & $Acc_H$ & $Acc_S$ & $Acc_N$ & $Acc_H$ \\
\midrule
ZS-CLIP~\cite{radford2021learning} & 49.59 & 58.10 & 53.51 & 51.44 & 72.40 & 60.15 & 50.52 & 65.25 & 56.83 \\
Tent~\cite{wang2020tent}          & 49.63 & 58.25 & 53.60 & 51.59 & 72.07 & 60.13 & 50.61 & 65.16 & 56.87 \\
TPT~\cite{shu2022test}            & 48.99 & 60.61 & 54.18 & 51.94 & 78.31 & 62.46 & 50.47 & 69.46 &  58.32 \\
DMN~\cite{zhang2024dual_cvpr}      & 51.37 & 61.87 & 56.13 & 49.87 & 67.83 & 57.48 & 50.62 & 64.85 & 56.81 \\
AdaNeg~\cite{zhang2024adaneg}      & 52.54 & 63.34 & 57.44 & 54.30 & 58.46 & 56.30 & 53.42 & 60.90 & 56.87 \\
OODD~\cite{yang2025oodd}          & 58.88 & 34.57 & 43.56 & 57.05 & 50.23 & 53.42 & 57.97 & 42.40 & 48.96 \\
AdaND~\cite{cao2025noisy}         & 62.47 & 27.17 & 37.87 & 59.81 & 57.43 & 58.60 & 61.14 & 42.30 & 48.24 \\
\rowcolor{gray!15} \textbf{DDE}   & \textbf{61.48} & \textbf{61.59} & \textbf{61.53} & \textbf{59.05} & \textbf{72.95} & \textbf{65.27} & \textbf{60.26} & \textbf{67.27} & \textbf{63.40} \\
\bottomrule
\end{tabular}
\end{table}

\begin{table}[t]
\caption{OOD detection results of zero-shot methods on the OpenOOD benchmark. ImageNet-1k is adopted as ID dataset.}\label{tab:openood}
\small
\setlength{\tabcolsep}{3pt}
\centering
\begin{tabular}{l|cc|cc}
\toprule
\multirow{2}{*}{ Methods } & \multicolumn{2}{|c|}{ FPR95 $\downarrow$} & \multicolumn{2}{|c}{$\mathrm{AUROC} \uparrow$} \\
\cline{2-5}
& Near-OOD & Far-OOD & Near-OOD & Far-OOD \\
\midrule
MCM~\cite{ming2022delving} & 79.02 & 68.54 & 60.11 & 84.77 \\
NegLabel~\cite{jiang2024negative} & 68.18 & 27.34 & 76.92 & 93.30 \\
AdaNeg~\cite{zhang2024adaneg} & 67.51 & 17.31 & 76.70 & 96.43 \\
AdaND~\cite{cao2025noisy} & 69.07 & 17.51 & 75.07 & 96.30 \\
\rowcolor{gray!15}
\textbf{DDE} & \textbf{58.64} & \textbf{15.58} & \textbf{84.52} & \textbf{96.74} \\
\bottomrule
\end{tabular}
\end{table}

\begin{table} [b]
\centering
\caption{Detailed OOD detection results of DDE on the OpenOOD benchmark, where ImageNet is adopted as the ID dataset.}
\label{tab:detailed_openood}
\footnotesize
\begin{tabular}{llcc}
\toprule
\textbf{Near-/Far-OOD} & \textbf{Datasets} & \textbf{FPR95 $\downarrow$} & \textbf{AUROC $\uparrow$} \\
\midrule
\multirow{3}{*}{Near-OOD} & SSB-hard  & 61.02 & 83.50 \\
 & NINCO  & 56.26 & 85.55 \\
 & Mean & 58.64 & 84.52 \\
\midrule
\multirow{4}{*}{Far-OOD} & iNaturalist  & 0.47 & 99.83 \\
 & Textures  & 14.42 & 96.78 \\
 & OpenImage-O  & 31.84 & 93.62 \\
 & Mean & 15.58 & 96.74 \\
\bottomrule
\end{tabular}
\end{table}

\noindent\textbf{Results of OpenOOD Benchmark.}
\cref{tab:openood} summarizes the zero-shot OOD detection performance on the comprehensive OpenOOD benchmark. Our proposed DDE achieves a new state-of-the-art across all evaluated settings.
Most significantly, in the Near-OOD scenario—which is widely regarded as the most challenging task due to the high semantic similarity between ID and OOD classes—DDE reduces the FPR95 from AdaNeg's 67.51\% to 58.64\% and substantially boosts the AUROC from 76.70\% to 84.52\%. This 7.82\% absolute improvement in AUROC underscores the effectiveness of our dual-denoising strategy in capturing precise class boundaries. In the Far-OOD setting, while existing methods like AdaNeg and AdaND already perform well, DDE further pushes the performance envelope, achieving the lowest FPR95 of 15.58\% and a superior AUROC of 96.74\%. These consistent gains across both Near- and Far-OOD benchmarks demonstrate that DDE is not only robust to test-time noise but also highly effective at distinguishing various out-of-distribution shifts.

\noindent\textbf{Detailed Results of Near-OOD Detection.}
~\cref{tab:detailed_openood} reports the dataset-specific performance of DDE. Our method demonstrates robust detection capabilities, achieving an average AUROC of 85.52\% on Near-OOD and 96.74\% on Far-OOD. Notably, DDE excels in the iNaturalist task with a near-perfect AUROC of 99.83\% and a minimal FPR95 of 0.47\%.

\noindent\textbf{Results of CIFAR10/CIFAR100}.
\cref{tab:cifar_results} shows the performance on CIFAR-10 and CIFAR-100 benchmarks. DDE consistently achieves the highest average $Acc_H$, reaching 91.58\% and 69.52\%, respectively. Compared to SoTTA, DDE provides a significant boost in both Seen ($Acc_S$) and Noisy ($Acc_N$) accuracy, demonstrating its superior robustness across diverse OOD types and varying task complexities.

\begin{table}[t]
\centering
\setlength{\tabcolsep}{2pt}
\caption{Zero-shot noisy TTA results on CIFAR-10 and CIFAR-100. \textbf{Bold} indicates the best performance.}
\label{tab:cifar_results}
\tiny
\begin{tabular}{ll *{15}{c}}
\toprule
\multirow{2}{*}{\textbf{ID}} & \multirow{2}{*}{\textbf{Method}} & \multicolumn{3}{c}{\textbf{SVHN}} & \multicolumn{3}{c}{\textbf{LSUN}} & \multicolumn{3}{c}{\textbf{Texture}} & \multicolumn{3}{c}{\textbf{Places}} & \multicolumn{3}{c}{\textbf{Avg}} \\
\cmidrule(lr){3-5} \cmidrule(lr){6-8} \cmidrule(lr){9-11} \cmidrule(lr){12-14} \cmidrule(lr){15-17}
& & $Acc_S$ & $Acc_N$ & $Acc_H$ & $Acc_S$ & $Acc_N$ & $Acc_H$ & $Acc_S$ & $Acc_N$ & $Acc_H$ & $Acc_S$ & $Acc_N$ & $Acc_H$ & $Acc_S$ & $Acc_N$ & $Acc_H$ \\
\midrule
\multirow{6}{*}{CIFAR10} 
& ZS-CLIP & 83.55 & 98.39 & 90.36 & 83.11 & 97.82 & 89.87 & 82.18 & 91.82 & 86.73 & 81.73 & 76.26 & 78.90 & 82.64 & 91.07 & 86.47 \\
& Tent    & 87.18 & 52.90 & 65.85 & 89.03 & 73.96 & 80.80 & 89.78 & 88.48 & 89.13 & 88.78 & 65.44 & 75.34 & 88.69 & 70.19 & 77.78 \\
& SoTTA   & 90.21 & 81.71 & 85.75 & 90.13 & 91.06 & 90.59 & 89.56 & 90.96 & 90.25 & 89.04 & 74.17 & 80.93 & 89.73 & 84.47 & 86.88 \\
& TPT     & 81.76 & 98.85 & 89.50 & 81.53 & 97.93 & 88.98 & 80.43 & 92.11 & 85.87 & 79.88 & 77.18 & 78.51 & 80.90 & 91.52 & 85.72 \\
\rowcolor{gray!15} & \textbf{DDE} & \textbf{91.50} & \textbf{98.89} & \textbf{95.05} & \textbf{91.71} & 93.82 & \textbf{92.75} & \textbf{90.23} & \textbf{93.43} & \textbf{91.80} & \textbf{90.45} & \textbf{82.58} & \textbf{86.34} & \textbf{90.97} & \textbf{92.18} & \textbf{91.58} \\
\midrule
\multirow{6}{*}{CIFAR100} 
& ZS-CLIP & 48.52 & 97.58 & 64.81 & 49.29 & 94.97 & 64.90 & 46.76 & 81.58 & 59.45 & 45.36 & 64.52 & 53.27 & 47.48 & 84.66 & 60.61 \\
& Tent    & 55.39 & 42.41 & 48.04 & 60.06 & 83.37 & 69.82 & 59.31 & 79.13 & 67.80 & 57.52 & 62.24 & 59.79 & 58.07 & 66.79 & 61.36 \\
& SoTTA   & 60.56 & 89.24 & 72.15 & 60.28 & 88.89 & 71.84 & 58.79 & 81.56 & 68.33 & 57.01 & 65.73 & 61.06 & 59.16 & 81.36 & 68.34 \\
& TPT     & 46.09 & 97.87 & 62.67 & 46.90 & 95.36 & 62.88 & 43.87 & 83.10 & 57.42 & 42.48 & 66.86 & 51.95 & 44.84 & 85.80 & 58.73 \\
\rowcolor{gray!15} & \textbf{DDE} & 58.51 & \textbf{98.46} & \textbf{73.40} & \textbf{60.60} & \textbf{89.40} & \textbf{72.24} & \textbf{61.26} & 82.76 & \textbf{70.41} & 58.40 & 66.12 & \textbf{62.02} & \textbf{59.69} & \textbf{84.19} & \textbf{69.52} \\
\bottomrule
\end{tabular}
\end{table}

\begin{figure*}[t]  
  \centering
  
  \begin{subfigure}[b]{0.24\linewidth}
    \centering
    \includegraphics[width=\linewidth]{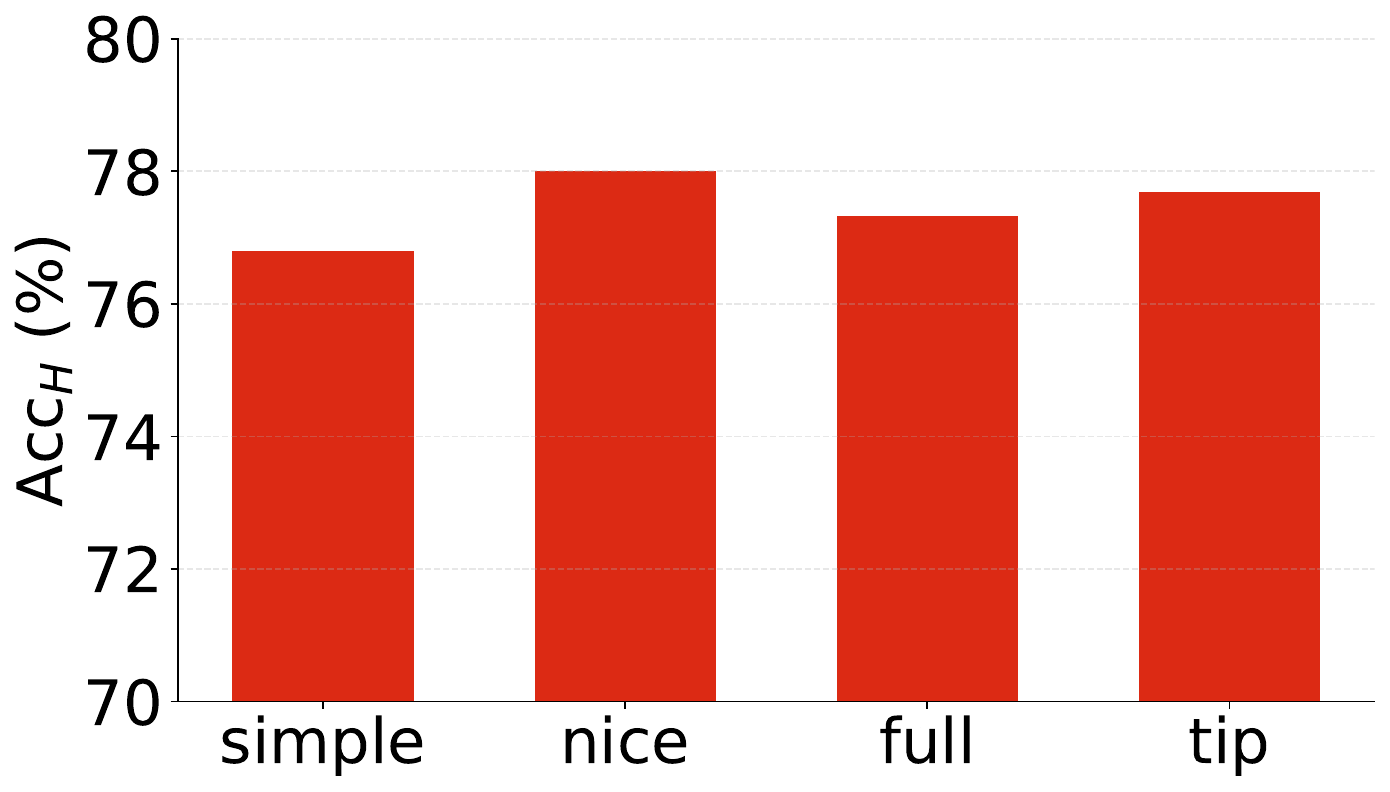}
    \caption{Prompt}
    \label{fig:ana_prompt}
  \end{subfigure}
  \hfill
  \begin{subfigure}[b]{0.24\linewidth}
    \centering
    \includegraphics[width=\linewidth]{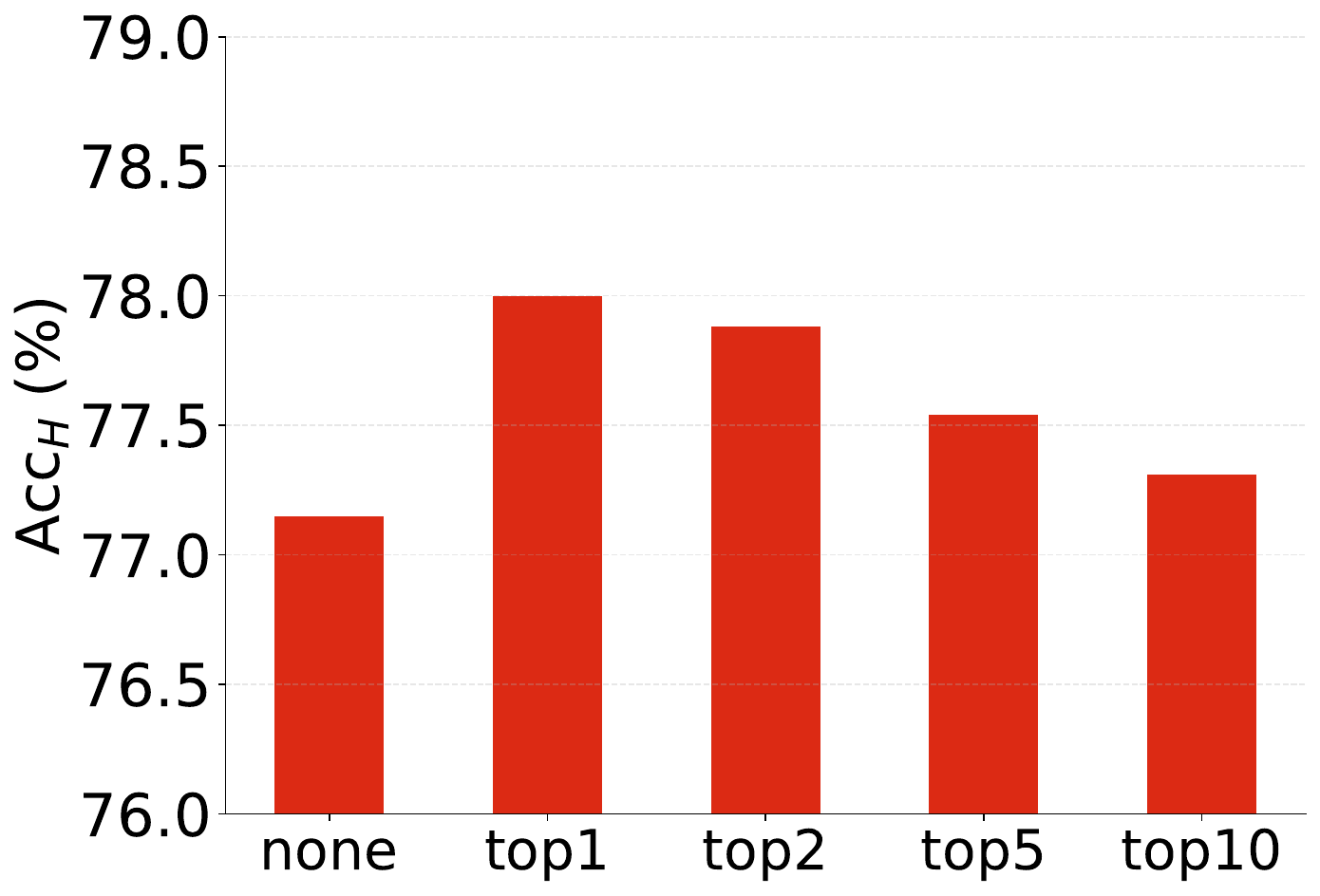}
    \caption{Exclusion $k$}
    \label{fig:ana_topk}
  \end{subfigure}
  \hfill
  \begin{subfigure}[b]{0.24\linewidth}
    \centering
    \includegraphics[width=\linewidth]{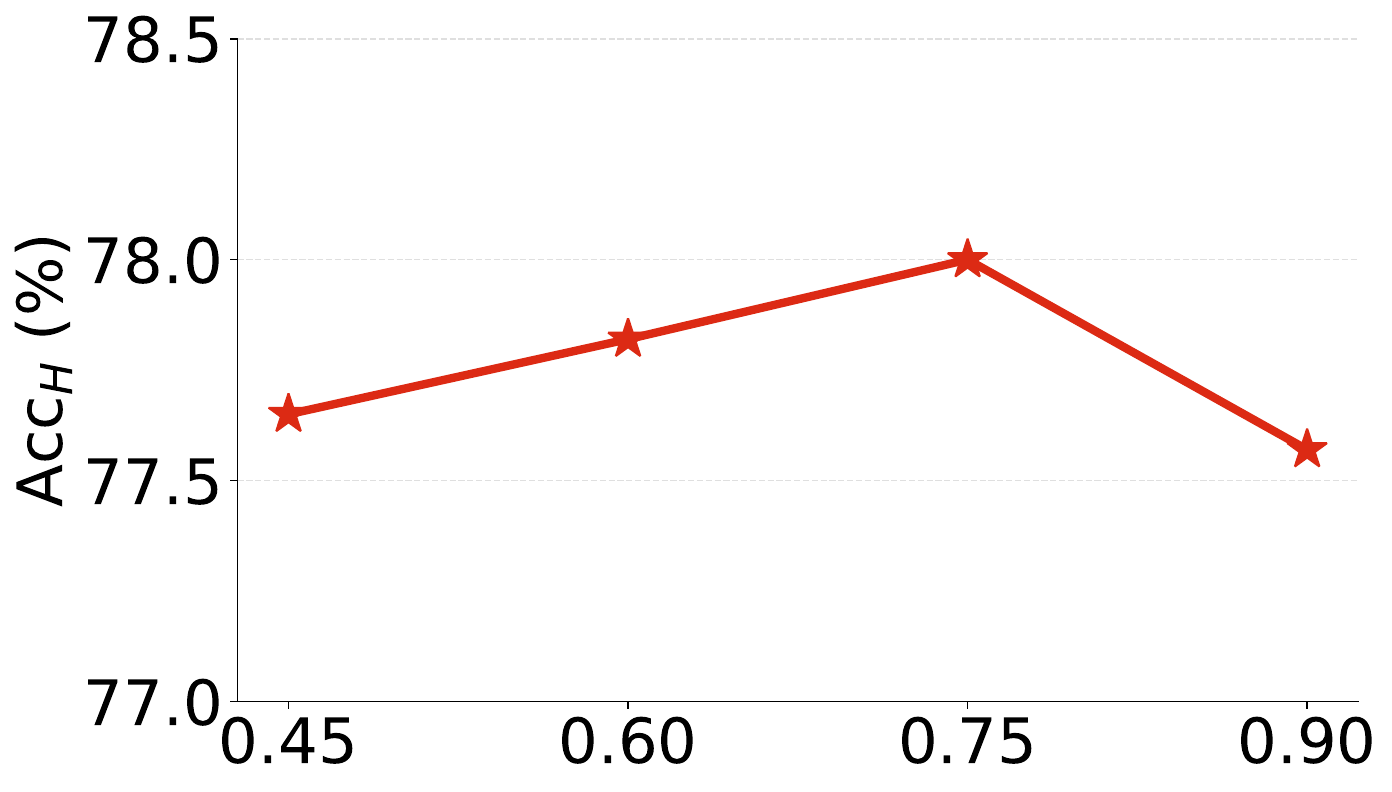}
    \caption{$\lambda_{pos}$}
    \label{fig:ana_pos}
  \end{subfigure}
  \hfill
  \begin{subfigure}[b]{0.24\linewidth}
    \centering
    \includegraphics[width=\linewidth]{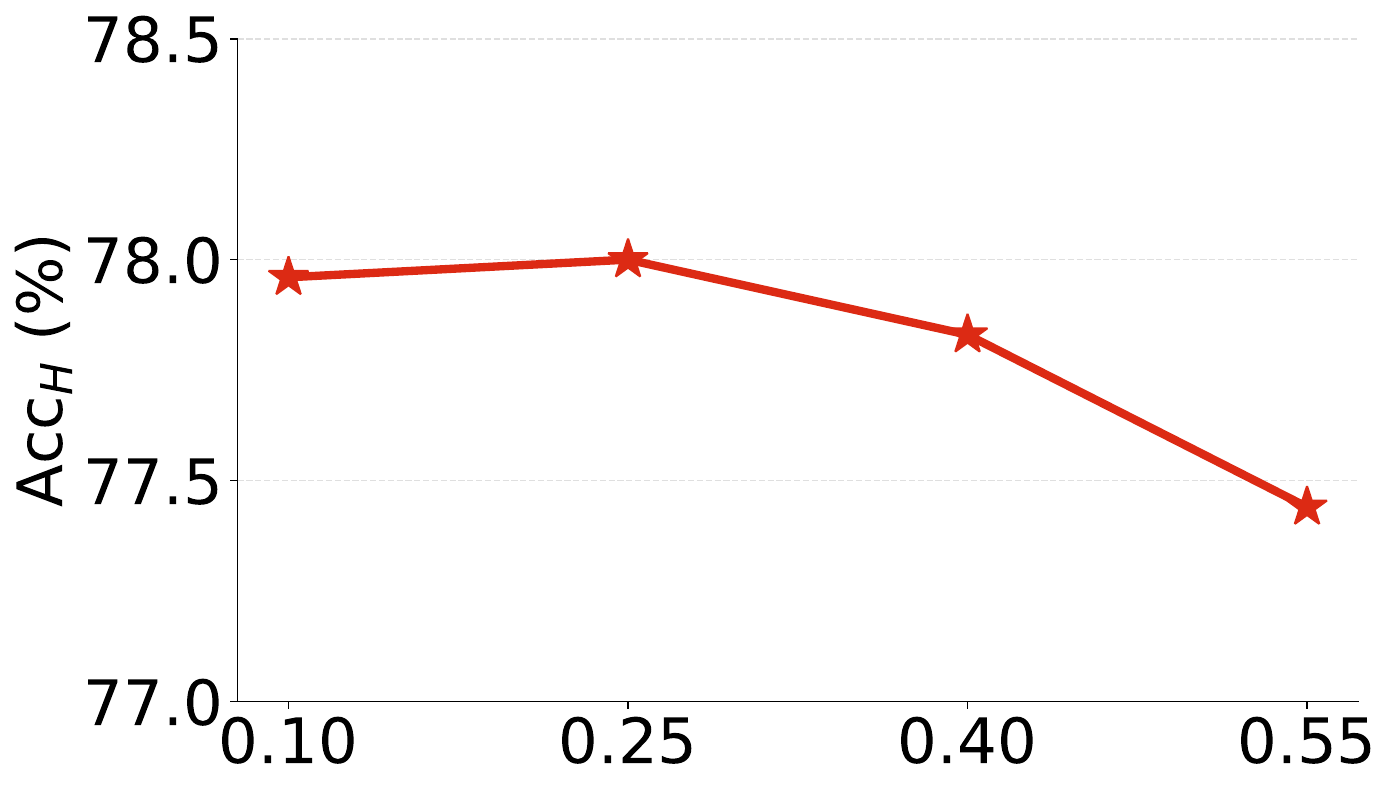}
    \caption{$\lambda_{neg}$}
    \label{fig:ana_neg}
  \end{subfigure}

  \caption{Sensitivity analysis of (a) prompt templates, (b) exclusion $k$, (c) positive threshold $\lambda_{pos}$, and (d) negative threshold $\lambda_{neg}$.}
  \label{fig:parameter_analysis}
\end{figure*}

\noindent\textbf{Robustness to Prompt Templates.}
~\cref{fig:ana_prompt} illustrates that our approach is relatively insensitive to prompt engineering, yielding consistent results across four different templates. While the `nice' template performs the best at 78.00\%, the `simple' baseline still achieves a high $\text{Acc}_H$ of 76.80\%. These findings demonstrate that the effectiveness of our method does not rely on meticulous prompt tuning.

\noindent\textbf{Sensitivity Analysis on Exclusion $TopK$.}
\cref{fig:ana_topk} illustrates the impact of the exclusion $k$ strategy on model performance. Compared to the baseline without exclusion (none, 77.15\%), incorporating this mechanism consistently improves the harmonic mean ($Acc_H$). The optimal performance is achieved at $k=1$ (78.00\%), which effectively filters out the most dominant noisy signals. As $k$ increases further, a slight performance degradation is observed, likely due to the over-exclusion of potentially useful semantic information.

\noindent\textbf{Hyperparameters Analysis.}
We further investigate the impact of positive and negative thresholds, $\lambda_{pos}$ and $\lambda_{neg}$, on the model's performance. As illustrated in \cref{fig:ana_pos}, the harmonic mean $Acc_H$ exhibits a "bell-shaped" trend with respect to $\lambda_{pos}$. The performance peaks at $\lambda_{pos} = 0.75$ (78.00\%), suggesting that an appropriately high threshold is essential for selecting high-confidence positive samples while avoiding the inclusion of excessive noise. Similarly, \cref{fig:ana_neg} shows the sensitivity of $\lambda_{neg}$, where the optimal result is achieved at $0.25$. A too low threshold may lead to the neglect of valuable negative constraints, while a too high threshold might introduce false negative signals, both resulting in a drop in accuracy. Overall, DDE maintains a relatively stable performance within a reasonable range of threshold values, demonstrating its robustness to hyper-parameters.


\end{document}